\documentclass{article} % For LaTeX2e

\usepackage{newtxtext}
\usepackage{amsmath,amssymb,amsthm}
\usepackage{booktabs}
\usepackage{longtable}
\usepackage{enumitem}          % tight contribution list in the introduction
\usepackage{graphicx}
\usepackage{microtype}
\usepackage{hyperref}
\usepackage{url}
\usepackage{fontawesome5}

\usepackage[accepted]{mlsys2025}
\makeatletter
\renewcommand{\mlsys@appearing}{\textit{Preprint.}}
\makeatother
\usepackage{titlesec}

\titlespacing*{\section}
  {0pt}{0.5ex plus 0.1ex minus 0.1ex}{0.25ex}

\titlespacing*{\subsection}
  {0pt}{0.4ex plus 0.1ex minus 0.1ex}{0.15ex}

\titlespacing*{\paragraph}
  {0pt}{0.2ex plus 0.1ex minus 0.1ex}{0.4em}
\setlist[itemize]{
    leftmargin=1.4em,
    itemsep=0pt,
    topsep=1pt,
    parsep=0pt,
    partopsep=0pt
}

\newcommand{\R}{\mathbb{R}}
\newcommand{\tile}{\mathrm{tile}}
\newcommand{\hada}{\odot}

\renewcommand{\algorithmiccomment}[1]{\hfill\scriptsize\texttt{// #1}}

\theoremstyle{plain}
\newtheorem{lemma}{Lemma}
\newtheorem{theorem}{Theorem}

\mlsystitlerunning{Turbo-GMM: GPU-Accelerated Gaussian Mixture Models}

\begin{document}

\twocolumn[
\mlsystitle{Turbo-GMM: FlashAttention-Inspired GPU-Accelerated Gaussian Mixture Model
}

\begin{mlsysauthorlist}
\mlsysauthor{Momin Abbas}{ibm}
\mlsysauthor{Sadia Asif}{rpi}
\mlsysauthor{Asad Ul Haq}{rpi}
\mlsysauthor{Kaoutar El Maghraoui}{ibm}
\end{mlsysauthorlist}

\mlsysaffiliation{ibm}{IBM Research}

\mlsysaffiliation{rpi}{Rensselaer Polytechnic Institute}

\mlsyscorrespondingauthor{Momin Abbas}{Momin.Abbas1@ibm.com}

\begin{center}
\faGithub\ \href{https://github.com/mominabbass/turbo-gmm}{github.com/mominabbass/turbo-gmm}
\end{center}

\mlsyskeywords{Gaussian Mixture Models, Expectation-Maximization, GPU Kernels, Kernel Fusion,
IO-Aware Algorithms, Flash Attention, Online Softmax, Memory-Bound Kernels, Triton, On-Chip Memory}

\begin{abstract}
Expectation-maximization (EM) for Gaussian mixture models (GMMs)
requires scoring $N$ data points against $K$ mixture components.
Prior fused GPU implementations, notably Flash-GMM, eliminate
materialization of the $N \times K$ responsibility matrix but
traverse the component dimension twice to compute normalization
terms and accumulate sufficient statistics. We introduce Turbo-GMM, a single-sweep fused EM kernel that
completes responsibility normalization and sufficient-statistic
accumulation within one traversal of the component dimension.
By retaining the complete component axis on chip, Turbo-GMM
avoids the second score computation required by the released
two-sweep Flash-GMM implementation. Its applicability is
limited by register pressure, occupancy, and the component
and feature dimensions, making it particularly suitable for
large datasets with relatively small numbers of components.
For larger configurations, Turbo-GMM employs a two-pass
fallback following the established fused design. We establish algebraic equivalence to standard EM in real arithmetic and evaluate sixteen configurations on an NVIDIA A100. Under the reported benchmarking protocol, Turbo-GMM achieves a median $2.15\times$ speedup over the evaluated Flash-GMM implementation and up to $5.3\times$ and $9.9\times$ over PyTorch and \texttt{scikit-learn}, respectively. The results demonstrate the potential of single-sweep component traversal while identifying on-chip resource constraints as the principal limitation of this approach.
\end{abstract}
]

\printAffiliationsAndNotice{}

% ============================================================================

\section{Introduction}

Gaussian mixture models (GMMs), commonly fitted using
expectation-maximization (EM), are widely used for density
estimation and soft clustering \citep{dempster1977em}.
Each EM iteration evaluates the likelihood of $N$ data
points under $K$ mixture components, normalizes these
scores into responsibilities, and accumulates sufficient
statistics for parameter updates. Efficient execution on
GPUs requires managing both the computational cost of
these operations and the movement of intermediate data.

Conventional implementations may materialize an
$N \times K$ responsibility matrix, introducing
substantial memory traffic and storage overhead.
Flash-GMM \citep{bloch2026flashgmm} addresses this
limitation by fusing the expectation and maximization
steps without materializing the responsibility matrix.
Consequently, eliminating this intermediate and its
associated memory overhead is an established benefit
of fused GMM implementations rather than a contribution
unique to Turbo-GMM.

An additional opportunity arises from the traversal
of the mixture components within a fused EM iteration.
The released Flash-GMM implementation processes the
component dimension twice: first to compute the
normalization terms and then to recompute component
scores and accumulate sufficient statistics.
Although the Flash-GMM paper also proposes a schedule
that retains input data across both traversals, its
released diagonal-covariance implementation reloads
the data during the second pass. Thus, reading the input
data once is not itself a new conceptual contribution.
The distinction we investigate is whether the second
traversal of the component dimension can be eliminated
altogether.

The challenge follows from a dependency between the
two EM reductions. Responsibility normalization requires
scores from all $K$ components for each data point,
whereas sufficient-statistic accumulation reduces
over the data-point dimension. A conventional
component-tiled fused implementation completes the
normalization before revisiting the component scores
to compute the statistics. However, when the entire
component dimension can be retained on chip alongside
a tile of data points, both reductions can be completed
during a single traversal. This eliminates the need
to recompute component scores in a second sweep.

Motivated by this observation, we introduce
\textbf{Turbo-GMM}, a single-sweep fused EM kernel
for diagonal-covariance GMMs. Turbo-GMM retains the
complete component dimension on chip, computes
responsibilities, and accumulates sufficient statistics
within the same tile visit. The kernel therefore
performs one component traversal and one read of the
input tile. Relative to two-sweep fused implementations,
its distinguishing improvement is the elimination of
the second component traversal and its associated
recomputation, rather than the elimination of the
responsibility matrix itself.
The single-sweep design requires on-chip state
proportional to $Kd$, making register pressure and
occupancy limiting factors as $K$ or $d$ increases.
On the evaluated A100, the dispatch rule selects this
kernel when $K_{\mathrm{pad}} \le 128$ and
$K_{\mathrm{pad}}d_{\mathrm{pad}} \le 4096$
(\S\ref{sec:dispatch}); these are implementation-specific
thresholds, not universal hardware limits. Otherwise,
Turbo-GMM uses a two-pass fused fallback.

We analyze the equivalence of the single-sweep and
two-pass formulations to standard EM in real arithmetic,
including arbitrary tilings and partial tiles
(Theorem~\ref{thm:exact}). We also characterize their
data-access costs under an explicit memory-traffic model
(Theorem~\ref{thm:traffic}). These analyses distinguish
the existing benefit of avoiding materialized
responsibilities from the additional savings in data
traversal and score recomputation enabled by the
single-sweep kernel.

We evaluate Turbo-GMM on sixteen configurations covering
different dataset sizes, component counts, and feature
dimensions using an NVIDIA A100 GPU. Under the reported
experimental protocol, Turbo-GMM achieves a median
$2.15\times$ speedup over the evaluated Flash-GMM
implementation, with improvements depending on whether
the component dimension can be retained efficiently
on chip. The results also identify cases in which
resource limitations make the two-pass fallback
necessary or preferable. The performance comparisons
are discussed in \S\ref{sec:results}, with limitations
in \S\ref{sec:discussion}.

\paragraph{Contributions.}
\begin{itemize}[leftmargin=1.4em, itemsep=0pt, topsep=1pt, parsep=0pt, partopsep=0pt]
    \item We develop a \textbf{single-sweep fused EM
    kernel} that completes responsibility normalization
    and sufficient-statistic accumulation within one
    traversal of the component dimension. Unlike
    existing two-sweep fused implementations, the
    kernel avoids recomputing component scores in a
    second traversal when the full component axis
    fits on chip (\S\ref{sec:flash}).

    \item We characterize the \textbf{resource constraints
    and operating regime} of single-sweep execution,
    including the effects of component count, feature
    dimension, register pressure, and occupancy.
    An empirically determined dispatch rule selects
    the single-sweep implementation where effective
    and a two-pass fallback otherwise
    (\S\ref{sec:dispatch}).

    \item We establish \textbf{algebraic correctness}
    in real arithmetic and analyze the data-access
    costs of both implementations, explicitly
    distinguishing inherited fusion benefits from
    the additional reductions associated with
    single-sweep execution (\S\ref{sec:exact}).

    \item We experimentally evaluate Turbo-GMM
    against fused GPU implementations and established
    GMM libraries, identifying the workloads in
    which single-sweep traversal improves latency
    and the configurations in which on-chip resource
    limitations restrict its benefits
    (\S\ref{sec:results}).
\end{itemize}
% ============================================================================
\section{Background and the Standard Implementation}
\label{sec:background}

In this section we establish the baseline that the rest of the paper measures against: the EM
iteration and our notation (\S\ref{sec:em}), the expanded score that libraries evaluate in practice
and the standard dataflow it yields (\S\ref{sec:lift}, Algorithm~\ref{alg:standard}), and the two
properties of that dataflow which decide whether the iteration can be fused at all
(\S\ref{sec:bottleneck})---the second of which \S\ref{sec:method} exploits.

\subsection{The EM Iteration}
\label{sec:em}

\paragraph{Data and notation.} Throughout, $X \in \R^{N \times d}$ holds $N$ data points as its rows
$x_i$, where $d$ is the dimension of each point. We denote by $\hada$ the elementwise product and by
$\oslash$ elementwise division, and let shapes broadcast in the usual way.
\paragraph{The mixture model.} We follow the standard parameterization of a Gaussian mixture
\citep{bishop2006pattern}. A mixture of $K$ diagonal Gaussians is described by three parameters:
mixture weights $\pi \in \R^{K}$, strictly positive and summing to one, which say how much of the
data each component accounts for; means $M \in \R^{K \times d}$ with rows $\mu_k$, one centre per
component; and variances $V \in \R_{>0}^{K \times d}$ with rows $\sigma^2_k$, one width per component
and per dimension. Positivity is what makes $\log \pi_k$ and $1 \oslash V$ well defined, and the
algorithm uses both. EM fits these parameters by ascending the log-likelihood
\vspace{-6pt}
\begin{equation}
  \ell(\pi, M, V) \;=\; \sum_{i=1}^{N} \log \sum_{k=1}^{K} \pi_k \,
  \mathcal{N}\!\left(x_i ; \mu_k, \operatorname{diag} \sigma^2_k\right)
  \label{eq:objective}
\end{equation}
\vspace{-6pt}
in two alternating stages.

\paragraph{The E-step.} The first stage asks how strongly each component explains each point. It
forms a score matrix $S \in \R^{N \times K}$ of log joint densities, then normalizes it along the
component axis:
\begin{align}
  S_{ik} &\;=\; \log \pi_k \;-\; \tfrac{d}{2}\log 2\pi
  \;-\; \tfrac{1}{2}\sum_{c=1}^{d} \log \sigma^2_{kc} \notag \\
  &\;-\; \tfrac{1}{2}\sum_{c=1}^{d} \frac{(x_{ic} - \mu_{kc})^2}{\sigma^2_{kc}} ,
  \label{eq:score}
\end{align}
\vspace{-6pt}
\begin{equation}
  L_i \;=\; \log \sum_{k=1}^{K} \exp S_{ik} , \qquad
  \Gamma_{ik} \;=\; \exp\!\left(S_{ik} - L_i\right) .
  \label{eq:estep}
\end{equation}
We call $L \in \R^{N}$ the row normalizers and $\Gamma \in \R^{N \times K}$ the
\emph{responsibilities}: $\Gamma_{ik}$ is the posterior probability that point $i$ was generated by
component $k$, so every row of $\Gamma$ sums to one. This is the matrix the paper is about.

\paragraph{The M-step.} The second stage refits each component to the points it now owns, weighting
every point by its responsibility. It reduces $\Gamma$ into three sufficient statistics, then
renormalizes them into parameters:
\begin{align}
  &n_k = \sum_{i=1}^{N} \Gamma_{ik}, \qquad
  s_k = \sum_{i=1}^{N} \Gamma_{ik}\, x_i, \notag \\
  &u_k = \sum_{i=1}^{N} \Gamma_{ik}\, x_i \hada x_i ,
  \label{eq:stats}
\end{align}
\begin{equation}
  \pi_k \leftarrow \frac{n_k}{N}, \qquad
  \mu_k \leftarrow \frac{s_k}{n_k}, \qquad
  \sigma^2_k \leftarrow \frac{u_k}{n_k} - \mu_k \hada \mu_k + \varepsilon ,
  \label{eq:mstep}
\end{equation}

Here $n_k$ is the soft number of points assigned to component $k$, while $s_k$ and $u_k$ are the
responsibility-weighted first and second moments of the data; $\mu_k$ on the right of
\eqref{eq:mstep} is the \emph{updated} mean, and we assume $n_k > 0$. The additive covariance
regularizer $\varepsilon \ge 0$ (\texttt{scikit-learn}'s \texttt{reg\_covar}) ensures strictly
positive variances when $\varepsilon > 0$ in real arithmetic. For $\varepsilon = 0$,
\eqref{eq:mstep} is the standard maximum-likelihood M-step, and EM does not decrease
\eqref{eq:objective} when the updates remain well defined \citep{wu1983convergence}. For $\varepsilon > 0$, the variance
update is regularized and is not the exact maximum-likelihood M-step; we make no monotonicity
claim for \eqref{eq:objective} or for an unspecified penalized objective. Since
$\sum_i L_i$ equals \eqref{eq:objective} at the incoming parameters, the E-step also provides
the log-likelihood used for convergence monitoring.

\subsection{The Score as a Single Bilinear Product}
\label{sec:lift}

Evaluated as written, \eqref{eq:score} sums $d$ elementwise terms for each of the $NK$ pairs, which
leaves idle the matrix-multiply units that supply most of a GPU's throughput. No library evaluates
it that way. Instead they expand the quadratic form and assemble the score from matrix products,
exactly as $k$-means expands $\|x_i - c_k\|^2_2$ to reuse norms and reach the same units \citep{johnson2021billion}. We state
the form we use below, since the fused kernels of \S\ref{sec:method} inherit it.

\begin{lemma}[Lift]
\label{lem:lift}
Let $\tau = 1 \oslash V \in \R^{K \times d}$ be the precisions, and define
\vspace{-6pt}
\begin{align}
  \tilde{X} &= \left[\, X \;\middle|\; X \hada X \,\right] \in \R^{N \times 2d},
  \notag \\
  \tilde{W} &= \left[\, M \hada \tau \;\middle|\; -\tfrac{1}{2}\tau \,\right]
  \in \R^{K \times 2d},
  \label{eq:lift}
\end{align}
% \vspace{-19pt}
\begin{align}
  b &\in \R^{K}, \qquad
  b_k \;=\; \log \pi_k \;-\; \tfrac{d}{2}\log 2\pi
  \;+\; \tfrac{1}{2}\sum_{c=1}^{d} \log \tau_{kc} \notag \\
  &\;-\; \tfrac{1}{2}\sum_{c=1}^{d} \mu_{kc}^2 \, \tau_{kc} .
  \label{eq:bias}
\end{align}
\vspace{-6pt}

Then the score matrix of \eqref{eq:score} satisfies exactly, with $\mathbf{1} \in \R^{N}$ the
all-ones column,
\vspace{-6pt}
\begin{equation}
  S \;=\; \tilde{X}\tilde{W}^{\top} + \mathbf{1}b^{\top} .
  \label{eq:bilinear}
\end{equation}
\vspace{-10pt}
\end{lemma}
The proof, by expanding the squared distance, is given in
Appendix~\ref{app:liftproof}.
% \begin{proof}
% With $\tau_{kc} = 1/\sigma^2_{kc}$,
% \begin{align*}
%   -\tfrac{1}{2}\sum_c (x_{ic}-\mu_{kc})^2 \tau_{kc}
%   &\;=\; \sum_c x_{ic}\,\mu_{kc}\tau_{kc}
%   \;-\; \tfrac{1}{2}\sum_c x_{ic}^2\,\tau_{kc} \\
%   &\;-\; \tfrac{1}{2}\sum_c \mu_{kc}^2 \tau_{kc} .
% \end{align*}
% The first two sums are the two blocks of $\tilde{X}\tilde{W}^{\top}$. The third is independent of
% $i$, so we put it in $b_k$ with $\log \pi_k$, the constant $-\frac{d}{2}\log 2\pi$, and the
% log-determinant $-\frac{1}{2}\sum_c \log \sigma^2_{kc} = +\frac{1}{2}\sum_c \log \tau_{kc}$.
% \end{proof}

In words, \eqref{eq:bilinear} says that the entire score matrix is one matrix product plus a
per-component offset. This is what libraries already compute: \texttt{scikit-learn} evaluates the
same form for diagonal covariances, as two matrix products against $X$ and $X \hada X$ plus a
per-component constant \citep{pedregosa2011scikit}, and Lemma~\ref{lem:lift} merely writes the two
products as one of width $2d$. Two consequences matter for a fused kernel. First, the score is
bilinear in
$(\tilde{X}, \tilde{W})$, so a tile of rows and the parameters determine a tile of $S$, which
lets a kernel build $S$ on chip. Second, the block $X \hada X$ depends on $x_i$ alone, so the
kernel forms it in registers and never stores it. Building $\tilde{W}$ and $b$ costs $\Theta(Kd)$
once per iteration, after which both stay resident. Algorithm~\ref{alg:standard} writes
\eqref{eq:score}-\eqref{eq:mstep} in these terms: five kernels handing $S$ and $\Gamma$ through
HBM.

Algorithm~\ref{alg:standard} is an archetype, not a transcription of one library: every count in
\S\ref{sec:bottleneck}, the four $N \times K$ reads of $S$ and $\Gamma$ included, prices
Algorithm~\ref{alg:standard} as written and measures no library. Its dataflow resembles the
array-API path of \texttt{scikit-learn} \citep{pedregosa2011scikit}, which keeps score and
log-responsibilities as two concurrent $N \times K$ arrays, forms $X \hada X$ as an
$N \times d$ temporary, and traverses $\Gamma$ once per statistic. cuVS
\citep{rapids2026cuvs} instead tiles the row axis and normalizes in place, bounding the live
responsibility block; \S\ref{sec:discussion} treats that stronger baseline.

\begin{algorithm}[tb]
\caption{Standard GMM-EM implementation (one iteration)}
\label{alg:standard}
\footnotesize
\begin{algorithmic}[1]
\REQUIRE Data $X \in \R^{N \times d}$ and parameters $\pi, M, V$ resident in HBM
\ENSURE Updated $(\pi_{\mathrm{new}}, M_{\mathrm{new}}, V_{\mathrm{new}})$, log-likelihood $\ell$
\STATE Precompute the lift $\tau \leftarrow 1 \oslash V$, $\tilde{W}$, $b$ by \eqref{eq:lift}-\eqref{eq:bias}
\STATE \texttt{// Stage 1: E-step}
\STATE \textbf{Kernel 1 (Score):} Load blocks of $X$ and of $\tilde{W}$ from HBM
\STATE Compute $S \leftarrow \tilde{X}\tilde{W}^{\top} + \mathbf{1}b^{\top}$ by \eqref{eq:bilinear}, as two products against $X$ and $X \hada X$
\STATE Write $S \in \R^{N \times K}$ to HBM \COMMENT{$NK$ floats committed to HBM}
\STATE \textbf{Kernel 2 (Normalize):}
\FORALL{row $i \in \{1, \dots, N\}$}
  \STATE Load $S_{i*}$ from HBM, compute $L_i = \log \sum_k \exp S_{ik}$ and $\Gamma_{i*} \leftarrow \exp(S_{i*} - L_i)$
\ENDFOR
\STATE Write responsibilities $\Gamma \in \R^{N \times K}$ to HBM \COMMENT{Second $N \times K$ round trip}
\STATE Write normalizers $L \in \R^{N}$ to HBM
\STATE \texttt{// Stage 2: M-step}
\STATE \textbf{Kernel 3 (Square):} Read $X$ from HBM
\STATE Write $X \hada X \in \R^{N \times d}$ to HBM \COMMENT{Recomputable from $X$ in registers}
\STATE \textbf{Kernel 4 (Accumulate):} Initialize $n \in \R^{K}$, $s, u \in \R^{K \times d}$ to $0$ in HBM
\STATE Read $\Gamma$, \textbf{gemv}: $n \leftarrow \Gamma^{\top} \mathbf{1}$ \COMMENT{1st read of $\Gamma$}
\STATE Read $\Gamma, X$, \textbf{gemm}: $s \leftarrow \Gamma^{\top} X$ \COMMENT{2nd read of $\Gamma$}
\STATE Read $\Gamma, X \hada X$, \textbf{gemm}: $u \leftarrow \Gamma^{\top} (X \hada X)$ \COMMENT{3rd read of $\Gamma$}
\STATE \textbf{Kernel 5 (Normalization):} Read $n, s, u$ from HBM
\STATE Apply the parameter updates \eqref{eq:mstep}
\STATE Write $(\pi_{\mathrm{new}}, M_{\mathrm{new}}, V_{\mathrm{new}})$ to HBM, and set $\ell = \sum_i L_i$
\end{algorithmic}
\end{algorithm}
\vspace{-8pt}

\subsection{What Governs Whether the Iteration Can Be Fused}
\label{sec:bottleneck}

Two properties of Algorithm~\ref{alg:standard} govern what a fused kernel can achieve. The first
says that removing the intermediate is worth doing at all; the second says how many times a kernel
that removes it must still read the data.
\vspace{-6pt}
\paragraph{A traffic ledger for Algorithm~\ref{alg:standard}.}
What follows counts elements moved by each kernel's own definition, not a profiled DRAM-counter
measurement (\S\ref{sec:exact} makes this explicit for the fused forms as well). Two $N \times K$
arrays dominate its HBM traffic: Kernel~1 writes $S$, Kernel~2 reads it and
writes $\Gamma$, and Kernel~4 reads $\Gamma$ once per statistic. The subtotal is $2NK$ elements
written and $4NK$ read, none of it output. On the input side $X$ is read three times, by Kernels~1,
3 and~4, and the $X \hada X$ temporary adds a round trip for a quantity Lemma~\ref{lem:lift}
recovers in one instruction, so the input side costs $5Nd$ against $\Theta(Kd)$ for the parameters
and the statistics. Once $K$ passes $5d/6$ (Theorem~\ref{thm:traffic}), an array no consumer inspects
prices the iteration, and Kernel~1 already runs at full throughput.
\citet{yang2026flashkmeans} argue this for the $k$-means distance matrix, and
\citet{bloch2026flashgmm} for the responsibilities.

\paragraph{The two reductions run along opposite axes.}
An EM iteration contains two reductions, and they are not alike. The statistics \eqref{eq:stats} sum
over the point index $i$, so they decompose over any partition of the rows and combine by ordinary
addition: a kernel may accumulate them from row blocks in any order. The normalizer
\eqref{eq:estep} sums over the component index $k$ instead, and admits no such partial commitment,
because $\Gamma_{ik}$ cannot be formed until the whole row $S_{i*}$ is known. The consequence is the
one the next section exploits. Since Lemma~\ref{lem:lift} already lets a kernel build a tile of $S$
on chip, one pass over the data suffices exactly when the component axis is resident there too, for
then the normalizer completes within a single tile visit and leaves the statistics nothing further to
read. That condition is what separates Algorithm~\ref{alg:flash} from Algorithm~\ref{alg:flash2}.

% ============================================================================
\vspace{-8pt}
\section{Method}
\label{sec:method}

In this section we introduce Turbo-GMM, a fused EM iteration that keeps the responsibility matrix on
chip and never gives it an address in memory. Which of two forms runs is decided by one property of
the problem shape: whether a whole component axis fits on chip beside a block of rows. When it does,
a single traversal of that axis suffices and the kernel reads the data exactly once
(\S\ref{sec:flash}); when it does not, a second traversal becomes unavoidable, and we fall back to
the design point of prior fused work (\S\ref{sec:flash2}). \S\ref{sec:dispatch} states the rule that
chooses between the two, and \S\ref{sec:exact} proves that both compute the iteration of
\S\ref{sec:em} exactly.

\subsection{Turbo-GMM: Single-Sweep Component Traversal}
\label{sec:flash}

Our goal is an iteration that reads $X$ once. The obstacle is the one \S\ref{sec:bottleneck}
identified: a responsibility cannot be used until its whole row of scores is known. Our way around
it is to arrange that the row is completed inside a single tile visit. Let $B_N$ denote the number of
rows a kernel instance holds on chip and $B_K$ the number of components it holds alongside them.
When $B_K \ge K$ the component axis is resident, so one visit sees an entire row of $S$, the
normalization of \eqref{eq:estep} finishes there, and the E-step and the M-step collapse into a
single kernel. We describe it formally in Algorithm~\ref{alg:flash}.

The kernel computes the score of Kernel~1 of Algorithm~\ref{alg:standard} arithmetically
unchanged---only its address differs---reads $X$ once, forms $X \hada X$ in registers, and writes out
nothing but the $\Theta(K + Kd)$ statistics.

\begin{algorithm}[tb]
\caption{Turbo-GMM, one-pass form (fused E-step and M-step, nothing materialized)}
\label{alg:flash}
\footnotesize
\begin{algorithmic}[1]
\REQUIRE $X \in \R^{N \times d}$, parameters $\pi, M, V$, tiles $B_N$ (points) and $B_K \ge K$ (components), so the component axis is resident; padded lanes predicated out
\ENSURE $(\pi_{\mathrm{new}}, M_{\mathrm{new}}, V_{\mathrm{new}})$, log-likelihood $\ell$
\STATE Precompute the lift $\tau \leftarrow 1 \oslash V$, $\tilde{W}$, $b$ by \eqref{eq:lift}--\eqref{eq:bias}
\STATE Initialize $n, s, u, \ell \leftarrow 0$ in HBM
\FORALL{point tile $X_{\tile}$ of size $B_N$ in parallel}
  \STATE Load $\tilde{W}, b$ into registers; initialize on-chip $\hat{n}, \hat{s}, \hat{u}, \hat{\ell} \leftarrow 0$
  \STATE Load $X_{\tile} \in \R^{B_N \times d}$ from HBM on chip \COMMENT{$X$ read exactly once}
  \STATE Form $X^{2}_{\tile} \leftarrow X_{\tile} \hada X_{\tile}$ in registers \COMMENT{$\tilde{X}$ never materialized}
  \STATE Compute $S \leftarrow \tilde{X}_{\tile}\tilde{W}^{\top} + \mathbf{1}b^{\top}$ on chip, $\tilde{X}_{\tile} = [\,X_{\tile} \mid X^{2}_{\tile}\,]$ \COMMENT{$S$ never in HBM}
  \STATE Per row $i$ of the tile: $m_i \leftarrow \max_k S_{ik}$, $\lambda_i \leftarrow \sum_k \exp(S_{ik} - m_i)$, $L_i \leftarrow m_i + \log \lambda_i$, $\Gamma_{i*} \leftarrow \exp(S_{i*} - m_i) \oslash \lambda_i$
  \STATE Accumulate \eqref{eq:stats}: $\hat{n} \mathrel{+}= \sum_i \Gamma$, $\hat{s} \mathrel{+}= \Gamma^{\top} X_{\tile}$, $\hat{u} \mathrel{+}= \Gamma^{\top} X^{2}_{\tile}$, $\hat{\ell} \mathrel{+}= \sum_i L_i$
  \STATE \textbf{atomic\_add} $(\hat{n}, \hat{s}, \hat{u}, \hat{\ell})$ into $(n, s, u, \ell)$ \COMMENT{additive: atomics legal}
\ENDFOR
\STATE Compute $(\pi_{\mathrm{new}}, M_{\mathrm{new}}, V_{\mathrm{new}})$ by \eqref{eq:mstep}
\end{algorithmic}
\end{algorithm}
% \vspace{-6pt}

% Two implementation details are easy to get wrong, and both bear on exactness.

\paragraph{Masking partial tiles.} When $K$ does not fill a tile, the padding columns must be masked
with a large \emph{finite} sentinel $-\beta$ ($\beta = 3 \times 10^{38}$ in our kernel, just inside
the \texttt{float32} range) rather than with $-\infty$, since $-\infty$ propagates
$-\infty - (-\infty) = \mathrm{NaN}$ into the reduction. The price is that in real arithmetic
$-\beta$ still contributes $\exp(-\beta - m_i) > 0$ to $\lambda_i$, vanishing only by
\texttt{float32} underflow.
Theorem~\ref{thm:exact} predicates padded lanes out exactly, so the kernel realizes it only up to
that underflow.

\paragraph{Reusing the exponentials.} Form $\Gamma$ as $\exp(S - m) \oslash \lambda$, which reuses
the exponentials already computed for $\lambda$, rather than as a second $\exp(S - L)$. Only the
one-pass form offers this reuse, because only there are those exponentials still on chip.

\paragraph{The ledger after fusion.}
Algorithm~\ref{alg:flash} gives neither $S$ nor $\Gamma$ an address, so both round trips counted in
\S\ref{sec:bottleneck} vanish, as does the $X \hada X$ temporary; all that remains proportional to
$N$ is the single read of $X$. Theorem~\ref{thm:traffic} states the three ledgers exactly, and
the saving over Algorithm~\ref{alg:standard} grows with $K/d$. Measured latency follows that
prediction in the $d$ direction but not in the $K$ direction, where our margin falls below parity from
$K = 512$ upward---the width at which the component axis stops fitting on chip, so that
Algorithm~\ref{alg:flash2} carries the comparison instead (\S\ref{sec:results}).

\subsection{The General Case: Two Passes When the Component Axis Does Not Fit}

When $K$ is too large for the component axis to stay resident, no single tile visit can complete
a row's normalizer, so Turbo-GMM falls back to a two-pass design at the point
\citet{bloch2026flashgmm} occupy: pass~1 sweeps the component axis once to accumulate only the row
normalizers, $\Theta(N)$ state rather than $\Theta(NK)$; pass~2 then re-reads the data once per
component tile, recomputes the score from the stored normalizer, and accumulates the sufficient
statistics. This trades one extra read of the data for covering every $K$, at the cost of the score
recomputation that Algorithm~\ref{alg:flash} avoids. Algorithm~\ref{alg:flash2}, including the
online-softmax merge rule it uses to combine partial normalizers exactly, is given in full in
Appendix~\ref{sec:flash2}.

% ============================================================================
\section{Exactness and Cost}
\label{sec:exact}

We establish that both fused kernels compute the
standard EM updates (Theorem~\ref{thm:exact}) and
characterize their HBM traffic (Theorem~\ref{thm:traffic}).
Proofs are provided in Appendix~\ref{app:proofs}.

\begin{theorem}[Exactness]
\label{thm:exact}
Fix $N, K \ge 1$, data $X \in \R^{N \times d}$ with $d \ge 1$, weights $\pi > 0$ and variances
$V > 0$; take $B_K \ge K$ in Algorithm~\ref{alg:flash}, so that its single component tile is the whole
axis, and assume padded lanes are predicated out on both axes, the statistic accumulators seeded at
$0$ and the fold of \eqref{eq:merge} seeded with any pair whose second coordinate is $0$. Partition
the rows
$\{1, \dots, N\}$ into nonempty tiles, and for Algorithm~\ref{alg:flash2} partition the components
$\{1, \dots, K\}$ likewise, each of its two passes choosing its partitions independently and pass~2
running after pass~1 has written $L$; tile sizes need divide neither $N$ nor $K$. Then, in real
arithmetic, Algorithms~\ref{alg:flash} and~\ref{alg:flash2} return the statistics $n$, $s$, $u$ of
\eqref{eq:stats} and the log-likelihood $\ell$ of \eqref{eq:objective} at the incoming
$(\pi, M, V)$, for every such partition and every order in which the tiles are visited. Moreover $n_k > 0$ for every $k$, so the updates
\eqref{eq:mstep} are defined and also agree.
\end{theorem}

Its content is that the two reductions of an iteration reassociate independently: the statistics
\eqref{eq:stats} fold over addition on the row axis, and the normalizer of \eqref{eq:estep} folds
over the $\oplus$ of \eqref{eq:merge} on the component axis. We
state exactness against \eqref{eq:objective}--\eqref{eq:stats} rather than against
Algorithm~\ref{alg:standard}, which is an archetype (\S\ref{sec:lift}) and so a weaker reference. The
theorem also chains across the iterations of a run: each row of $\Gamma$ sums to one, so
$\sum_k n_k = N$ and the updated weights again lie in the open simplex, while
$u_k \oslash n_k - \mu_k \hada \mu_k$ is a weighted variance and hence nonnegative, so
$V_{\mathrm{new}} \ge \varepsilon$ and the output of one iteration satisfies the hypotheses of the
next whenever $\varepsilon > 0$ (Appendix~\ref{app:proofs}).

\begin{theorem}[HBM traffic]
\label{thm:traffic}
Count elements moved between HBM and the chip, charging one for a read, one for a write, and two for
an atomic update, with no cache between the two; on-chip state is free. Let $T_N = \lceil N / B_N \rceil$ and
$T_K = \lceil K / B_K \rceil$ be the tile counts. In Algorithm~\ref{alg:flash} the $T_N$ row tiles are
distributed over $P \le T_N$ programs, each loading the lift once and committing its partial
statistics once; this is the launch our kernel uses and the only place the mapping enters a ledger.
Then one iteration moves
\begin{align}
  &\underbrace{5Nd + 6NK + \Theta(N + Kd)}_{\textnormal{Algorithm~\ref{alg:standard}}}, \notag \\
  &\underbrace{Nd + \Theta(PKd)}_{\textnormal{Algorithm~\ref{alg:flash}}}, \notag \\
  &\underbrace{(1 + T_K)(Nd + N) + \Theta(T_N K d)}_{\textnormal{Algorithm~\ref{alg:flash2}}}
  \label{eq:ledgers}
\end{align}
elements, the parameter terms $\Theta(PKd)$ and $\Theta(T_N Kd)$ being upper bounds, since the lift is
only $2Kd + K$ elements and may in practice stay resident in a last-level cache across row tiles.
Whenever $PK = O(N + K)$, Algorithm~\ref{alg:flash} moves $\Theta(Nd + Kd)$, which is
optimal up to constants among exact algorithms that receive $X$ and return
$(\pi_{\mathrm{new}}, M_{\mathrm{new}}, V_{\mathrm{new}})$ in HBM, while
Algorithm~\ref{alg:standard} moves $\Theta(Nd + NK)$ whenever $\min(K, d) = O(N)$, in particular
whenever $K \le N$.
\end{theorem}

The full term-by-term accounting of each ledger, including why the two-pass
form's refetched lift term dominates its reads of $X$, is in Appendix~\ref{app:proofs}.

Our verification is stronger than the theorem needs: it compares against the \emph{centered}
\eqref{eq:score}, which also covers Lemma~\ref{lem:lift}. That lemma is an exact algebraic identity,
true by direct expansion for every $d$ (Appendix~\ref{app:proofs}). We check both tiled forms
against a $60$-decimal-digit \texttt{mpmath} extended-precision reference \citep{mpmath} on a battery of small cases, including
$K = 1$ and $K > N$, at both of our numeric paths (\texttt{ieee} and \texttt{tf32x3}, three
tensor-core passes at reduced input precision with an \texttt{fp32} accumulator, against one pass
for \texttt{bf16}), and against
\texttt{scikit-learn}'s and PyTorch's own E- and M-step primitives; separately, we sweep a
double-precision materializing oracle over multiple larger shapes and tile sizes. Across both
checks the tiled algorithms agree with the untiled computation to within ordinary floating-point
roundoff, not a difference in the underlying mathematics. In \texttt{float64} the summation order
differs, so agreement there is a relative bound rather than a bitwise one.
% \vspace{-6pt}
\paragraph{What the theorem does not cover.} Theorem~\ref{thm:exact} and the checks above concern
real arithmetic, and the \texttt{float32} kernel departs from its hypotheses twice. It masks padded
columns with the finite sentinel of \S\ref{sec:flash} rather than predicating them away, so it
realizes the theorem only inside the single-precision exponent range; and where a well-separated
component underflows $\Gamma_{ik}$, and $n_k$ with it, it divides by $n_k$ plus ten \texttt{float64}
machine epsilons, returning \eqref{eq:mstep} of a perturbed argument. We check it by measurement
instead (\S\ref{sec:results}, \S\ref{sec:ablation}).

\paragraph{A numerical precondition inherited from the expansion.}
The lifted score subtracts terms of magnitude $\|x\|^2 \tau$, losing relative accuracy as the
data moves away from the origin, so implementations must center the data
(\S\ref{sec:discussion}); Appendix~\ref{app:proofs} quantifies the loss and traces it to the same
cancellation that destabilizes a naive sum-of-squares variance formula \citep{chan1983algorithms}.

% ============================================================================

\section{Experiments}
\label{sec:experiments}

\begin{figure*}[t]
\centering
\includegraphics[width=\textwidth]{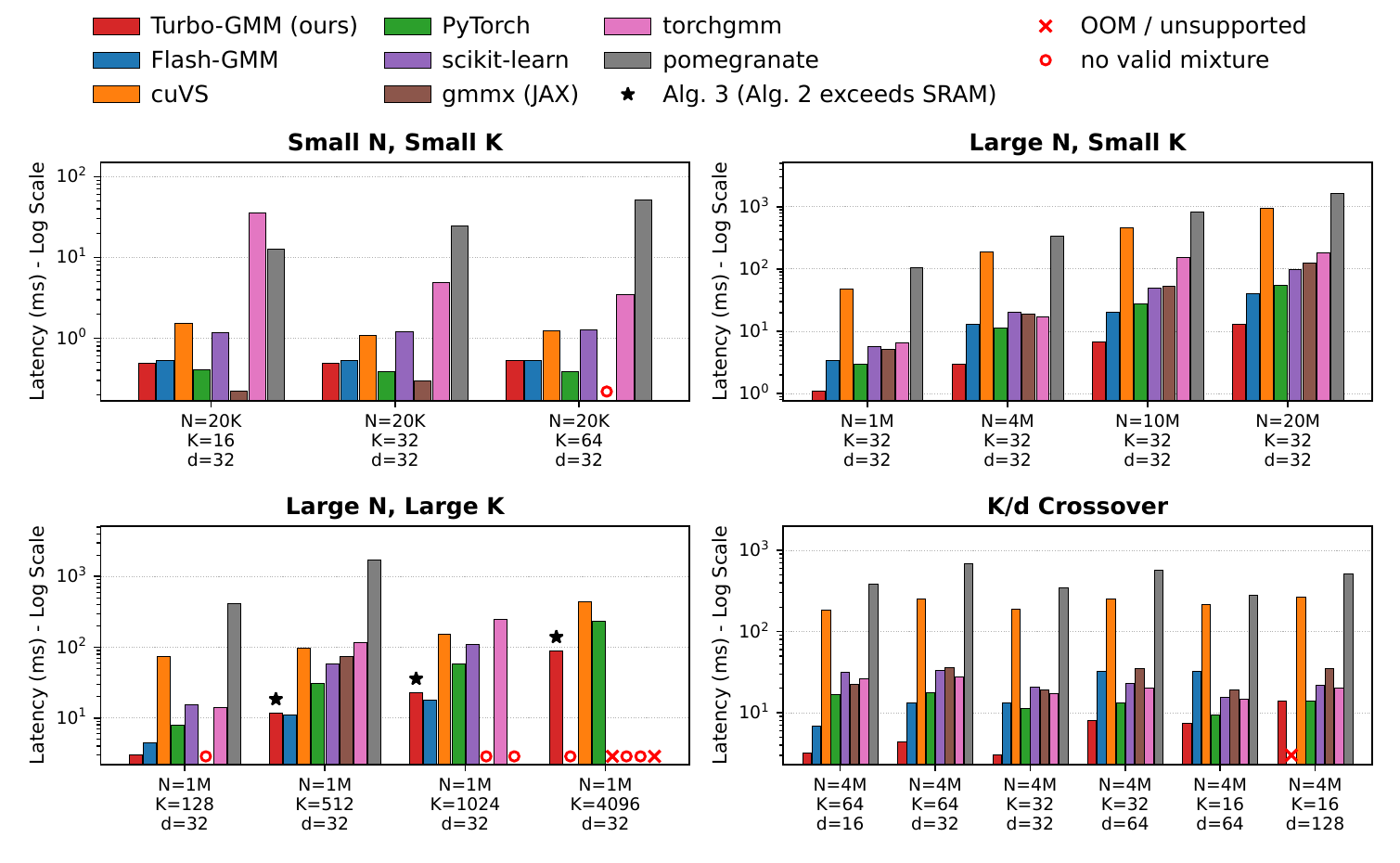}
\caption[Latency per EM iteration against seven baselines.]%
{\textbf{Latency per EM iteration} against seven baselines on one NVIDIA
A100-SXM4-80GB, four regimes, logarithmic axis, lower is better. \textbf{A star
marks the three configurations that fall back to Algorithm~\ref{alg:flash2}
because Algorithm~\ref{alg:flash} does not fit on chip}; Algorithm~\ref{alg:flash}
runs elsewhere. Axis-floor markers returned no valid fit, keyed in the legend; one such point,
Flash-GMM at $K = 4096$, is nonetheless discussed by latency in \S\ref{sec:results} and
Appendix~\ref{app:relation} as a disclosed exception to this criterion, so it is absent from this
figure but present in those aggregate numbers. Protocol and validity criterion: \S\ref{sec:setup};
dispatch: \S\ref{sec:dispatch}.}
\label{fig:latency}
\end{figure*}

Theorem~\ref{thm:exact} says the fused iteration computes the right thing and
Theorem~\ref{thm:traffic} that it moves less data. Neither says it runs faster: fusion buys the
traffic it saves with on-chip state, and in the two-pass form with a recomputed score. What the
experiments locate is the boundary of that trade over $(N, K, d)$. We evaluate three claims.

\begin{itemize}[leftmargin=1.4em, itemsep=0pt, topsep=1pt, parsep=0pt, partopsep=0pt]
  \item \textbf{Latency.} Against the closest fused kernel, Turbo-GMM is faster on eleven of the
        fifteen shapes both complete, by a median $2.15\times$; against the libraries a
        practitioner would reach for, by up to $5.3\times$ over PyTorch and $9.9\times$ over
        \texttt{scikit-learn}. The advantage is conditional, and we identify the three conditions
        that remove it.
  \item \textbf{Memory.} Peak memory is flat in the number of components---$128$ to $139$~MB from
        $K = 128$ to $4096$, against $3.2$ to $47$~GB for implementations that materialize the
        responsibilities. This claim carries no condition, and at the widest setting it decides
        feasibility rather than speed.
  \item \textbf{Where the gain comes from.} An ablation separates the single sweep of the component
        axis, the one change without precedent, from the two fusions that precede it. The single
        sweep is worth up to $19.4\times$ over our own two-pass fallback (not a margin over
        Flash-GMM; the comparable Flash-GMM control point is $2.07\times$, Appendix~\ref{app:relation}),
        and at $d = 128$ it is the only one of the three that keeps fusion profitable at all.
\end{itemize}

Latency is always reported at fixed quality: we re-score every arm's returned parameters in double
precision and admit only valid fits. For the full protocol, the dispatch measurements, the ablation
ladder and the cells that failed, see Appendix~\ref{app:details}, whose Table~\ref{tab:fullresults}
gives every arm's latency, memory and gate outcome at all sixteen shapes; the comparison against
prior fused work has its own Appendix~\ref{app:relation}.

\subsection{Setup}
\label{sec:setup}
\paragraph{Baselines.} Figure~\ref{fig:latency} compares
Turbo-GMM against seven baselines. Flash-GMM
\citep{bloch2026flashgmm} is the closest prior work:
it fuses the same iteration but traverses the component
axis twice, isolating the contribution of our single
sweep (\S\ref{sec:flash2}). cuVS
\citep{rapids2026cuvs} is the strongest library baseline
and the only other row-tiled implementation. The
remaining libraries are PyTorch \citep{paszke2019pytorch},
\texttt{scikit-learn} \citep{pedregosa2011scikit},
\texttt{gmmx} \citep{donath2024gmmx},
\texttt{torchgmm} \citep{varrone2023torchgmm}, and
\texttt{pomegranate} \citep{schreiber2018pomegranate};
versions are listed in Appendix~\ref{app:impl}.
% \vspace{-6pt}
\paragraph{Protocol.} Every arm is diagonal-covariance and starts from one seeded dataset and one
shared initialization, on one NVIDIA A100-SXM4-80GB. The sweep covers sixteen shapes $(N, K, d)$
in four regimes---large $N$, large $K$, a $K/d$ sweep, and small $N$---one of which belongs to two,
so Figure~\ref{fig:latency} shows seventeen bar groups. Both fused kernels run at \texttt{tf32x3},
which makes the comparison between them one of dataflow rather than of number format; the libraries
run at \texttt{fp32}---for PyTorch specifically this means explicitly disabling its
\texttt{allow\_tf32} matmul setting, an exposed control we turn off rather than a default the
library lacks. Two choices run
against us, and we flag them here rather than in an appendix: the latency estimator is not
identical across arms, mildly favouring cuVS, and we widened Flash-GMM's published autotune list so
that it could select the small blocks \texttt{fp32} needs, cutting our largest margin from
$4.37\times$ to $2.46\times$ (Appendix~\ref{app:details}).

\paragraph{What counts as a result.} A latency means nothing without the objective the arm reached,
and some libraries return an unusable model without saying so. We therefore re-score every arm's
returned parameters in double precision and accept a fit only if its log-likelihood is finite and,
for arms that return mixture weights, $\sum_k \pi_k = 1$. Runs that fail the gate sit at the axis
floor of Figure~\ref{fig:latency} and drop out of every memory and aggregate-count comparison
below, with one disclosed exception: \S\ref{sec:results} and Appendix~\ref{app:relation} keep and
explicitly flag Flash-GMM's $K = 4096$ latency point even though it fails this gate, since a single
E-step there is itself exact (Appendix~\ref{app:ibm}) and we would rather disclose a borderline
inclusion than silently drop a data point that completed.

\subsection{Which algorithm runs where}
\label{sec:dispatch}

Algorithm~\ref{alg:flash} runs wherever the component axis fits on chip and
Algorithm~\ref{alg:flash2} only where it does not, so before reading any latency we have to fix
which of the two produced it. Two limits bound the fit. The first is capacity: the loop-carried
state $(\hat n, \hat s, \hat u)$ spans the whole component axis, $\Theta(Kd)$, and past
$K_{\mathrm{pad}} d_{\mathrm{pad}} \approx 10^4$ the compiler cannot place it in the NVIDIA A100's
shared memory at all. The second is occupancy, and it binds first: the accumulator spills registers
well inside the capacity bound, and the cliff does not follow $K_{\mathrm{pad}} d_{\mathrm{pad}}$.
Dispatching on capacity alone would therefore be wrong, and measurably so: holding the footprint
fixed and trading $K$ against $d$ swings the ratio between the two algorithms by $250\times$, and
one configuration that capacity admits costs the one-pass form $17\times$. We consequently dispatch
on two measured conditions rather than one derived one, taking Algorithm~\ref{alg:flash} when
$K_{\mathrm{pad}} \le 128$ and $K_{\mathrm{pad}} d_{\mathrm{pad}} \le 4096$. For the capacity
arithmetic and the probes behind the rule, see Appendix~\ref{app:details}.

Under that rule three of the sixteen shapes, $K \in \{512, 1024, 4096\}$ at $N = 10^6$, fall to
Algorithm~\ref{alg:flash2} and carry a star in Figure~\ref{fig:latency}. On the other thirteen,
where both of our algorithms run, the one-pass form wins every shape---median $2.1\times$ over our own two-pass fallback, and
$19.5\times$ over it at $d = 128$, one measured point rather than a curve in $d$---because it sweeps
the component axis once and never recomputes the score. (This is a margin over our own fallback, not
over Flash-GMM; \S\ref{sec:results} and Appendix~\ref{app:relation} give the margins over Flash-GMM.)

\subsection{Latency and memory}
\label{sec:results}

\paragraph{All arms reach the same optimum.} Where two or more arms produced a valid fit, they
agree to $2.2 \times 10^{-6}$ relative on the final log-likelihood and to $1.1 \times 10^{-3}$ on
the adjusted Rand index of the induced clustering, and the two fused kernels agree to $4.4 \times
10^{-7}$. What follows compares cost at fixed quality.

\paragraph{Peak memory is flat in $K$, and at the widest setting it decides feasibility.}
Turbo-GMM never writes the $N \times K$ responsibilities, so its inputs govern its peak: from
$K = 128$ to $K = 4096$ at $N = 10^6$ that peak grows only from $128$ to $139$~MB, a factor of
$1.1$, against $3.2$ to $47$~GB for the largest baseline peak with a valid fit---$25\times$ to
$338\times$ ours, over the four library arms that report peak memory at all. The consequence is
not a constant factor. At $K = 4096$, \texttt{scikit-learn} exhausts the $80$~GB device and
\texttt{pomegranate} fails outright, while Turbo-GMM completes in $139$~MB.
% \vspace{-6pt}
\paragraph{Against libraries, the advantage is ordered in $d$ and disappears at small $N$.}
Measured against the fastest library with a valid fit at each shape, Turbo-GMM is
$2.7$--$4.2\times$ faster at large $N$ and $2.5$--$2.6\times$ at large $K$; taken one library at a
time the maxima are $5.3\times$ over PyTorch and $9.9\times$ over \texttt{scikit-learn}. Across the
$K/d$ sweep the range is wider, $0.99$--$5.3\times$, and it falls in $d$ at every fixed $K$ we
measure, reaching parity at $K = 16$, $d = 128$: this is the shape at which the $\Theta(NK)$
intermediate we remove drops below the input traffic that no implementation avoids, so a traffic
argument no longer separates us (Theorem~\ref{thm:traffic}). At $N = 20\,000$ we are slower still,
$0.45$--$0.73\times$ the best baseline, because the responsibility matrix is too small there for
its removal to pay against launch and reduction overheads. Fit a small mixture with a library.

\paragraph{Against the closest fused kernel, the comparison splits at the on-chip boundary.}
Flash-GMM times fifteen of the sixteen shapes, failing to launch at $d = 128$ on its published
list. Turbo-GMM, dispatched by \S\ref{sec:dispatch}, wins eleven of the fifteen by $1.06\times$ to
$4.28\times$---median $2.15\times$ over all fifteen and $3.02\times$ over the wins. The four
exceptions are placed rather than scattered. One is the small-$N$ tie ($0.99\times$). The other
three are exactly the shapes at which Algorithm~\ref{alg:flash} has left the chip and our two-pass
fallback meets their kernel at its own design point, and we lose all three: $0.93\times$,
$0.78\times$ and $0.73\times$ at $K = 512$, $1024$ and $4096$. The margin is thus ordered in $K$
only through the dispatch boundary: at $d = 32$ it falls monotonically as $K$ rises and reverses
exactly where the fallback takes over (Appendix~\ref{app:relation}). We have not identified the
mechanism of that reversal, and Theorem~\ref{thm:traffic} does not supply one, since it separates
the two designs on no term (\S\ref{sec:discussion}). One caveat at
$K = 4096$: their kernel completes in $65.2$~ms but returns mixture weights off the simplex
($\sum_k \pi_k = 4.3 \times 10^3$), traced in Appendix~\ref{app:ibm} to a collapsed component
variance, so its latency counts against a fit our own gate rejects.
Appendix~\ref{app:relation} places the two designs side by side.
% \vspace{-6pt}
\paragraph{Three libraries fail silently.} The gate rejected six otherwise-successful runs, spread
over \texttt{gmmx}, \texttt{torchgmm} and \texttt{pomegranate}, which completed, reported success,
and returned a non-finite log-likelihood. Four of the six sit at $K \ge 1024$, the regime this paper
is about, so a harness that trusted exit status rather than re-scoring the returned parameters would
enter all six in its table (Appendix~\ref{app:details}). Turbo-GMM returned a valid fit at all
sixteen shapes.
% \vspace{-6pt}
\subsection{Where the gain comes from}
\label{sec:ablation}

\begin{figure*}[t]
\centering
\includegraphics[width=\textwidth]{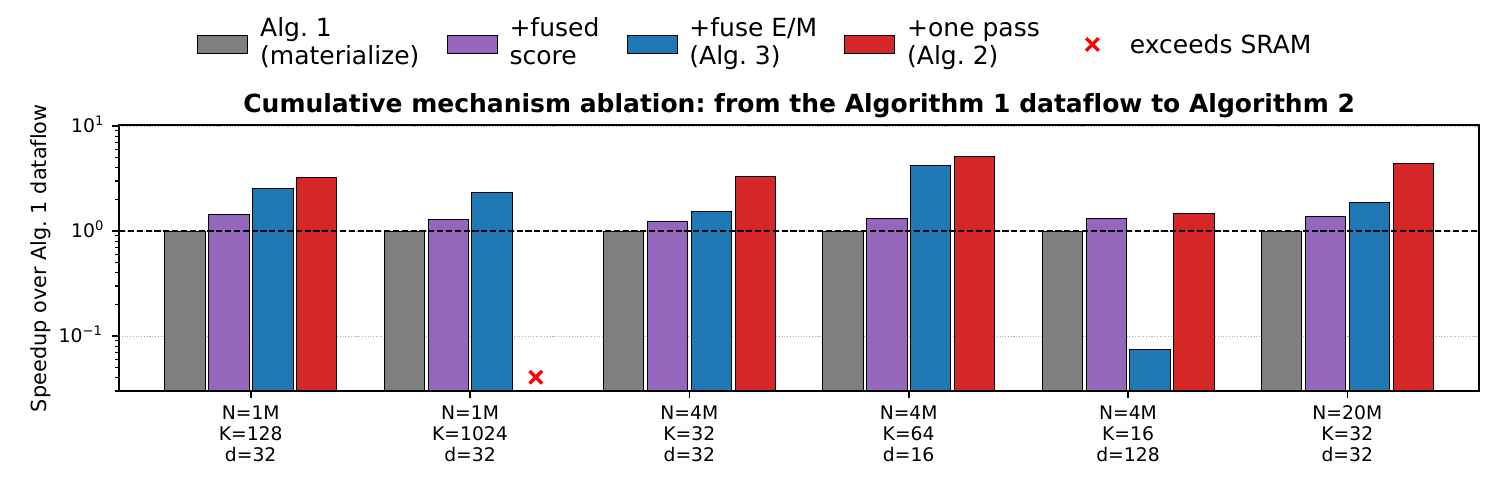}
\caption[Cumulative mechanism ablation.]%
{\textbf{Cumulative mechanism ablation.} Each bar adds one dataflow change to the ones below it;
height is speedup over the Algorithm~\ref{alg:standard} dataflow, so the bars multiply and the
dashed line is parity. \textbf{The middle rung is not monotone}: at $d = 128$ our two-pass
Algorithm~\ref{alg:flash2} runs $17\times$ slower than not fusing, and only the one-pass
Algorithm~\ref{alg:flash} recovers it---opposite in direction to the $17\times$ of
\S\ref{sec:dispatch}, a loss of Algorithm~\ref{alg:flash} at $K = 256$, $d = 16$. A red $\times$
marks a rung that exceeds SRAM. Every rung agrees with the reference to $5 \times 10^{-6}$; see
\S\ref{sec:ablation}.}
\label{fig:ablation}
\end{figure*}

Figure~\ref{fig:latency} says where the fused iteration is faster; it does not say which change is
responsible, and the changes are not equally novel. Three separate Algorithm~\ref{alg:standard}
from Algorithm~\ref{alg:flash}: fusing the score into one kernel, which any library could adopt;
fusing the E- and M-steps to remove the $N \times K$ intermediate, which \citet{bloch2026flashgmm}
also do; and the single sweep of the component axis, which has no precedent. If the first two
carried the margin, the advice this paper gives would stop at a fused two-pass kernel. We therefore
build a cumulative ladder, each rung adding one change to the ones below it so that the factors
compose, and run all four rungs on six configurations from Figure~\ref{fig:latency}. Every rung
agrees with the baseline to $5 \times 10^{-6}$ relative on all four returned quantities, so the
ladder varies dataflow and not arithmetic. For the rungs, the protocol and two caveats, see
Appendix~\ref{app:details}.
\vspace{-6pt}
\paragraph{The three factors are unequal, and which of them dominates depends on the shape.} The
score kernel returns $1.25$--$1.44\times$; the E/M fusion $1.24$--$3.18\times$ where it helps at
all; the single sweep $1.21$--$2.32\times$ at $d \le 32$. Where all four rungs run---five
configurations---the product ranges from $1.46\times$ to $5.14\times$. What licenses reading the
rest is that the first rung is also the most tightly bounded: the one rung that confounds dataflow
with implementation language never exceeds $1.44\times$, so it cannot account for a product of
$5.14\times$.
\vspace{-10pt}
\paragraph{At $d = 128$ the single sweep is not an optimization but the reason fusion pays at all.}
The middle rung is not monotone, and this is the central result of the ablation. At $N = 4 \times
10^6$, $K = 16$, $d = 128$, our two-pass Algorithm~\ref{alg:flash2} runs $17.4\times$ slower than
leaving the steps apart and $13.3\times$ slower than the materializing baseline; the single sweep
then recovers $19.4\times$ relative to our own two-pass form (not a margin over Flash-GMM), a net
$1.46\times$ over the standard dataflow, re-measuring the $19.5\times$ of \S\ref{sec:dispatch}
(also vs.\ our own two-pass form) to within $1\%$. The mechanism is the score recomputation
that a second sweep forces. Two limits bound how far this generalizes. A control at the same shape
charges most of the $17.4\times$ to our implementation rather than to two-sweep fusion as such---at
this shape, Flash-GMM (an independently written two-sweep kernel) regresses by only $2.07\times$
against the materializing baseline, not by an order of magnitude (Appendix~\ref{app:relation}): this
is Flash-GMM's own number at $d = 128$, the one to use for competing published work, not the
internal $19.4\times$ above. And $d = 128$ is the widest configuration we measure and the only one
at which the inversion appears, so this is an existence result rather than a curve in $d$.
\vspace{-6pt}
\paragraph{Latency and footprint come from different rungs.} The ladder also localizes the memory
result of \S\ref{sec:results}: at every shape but the widest $d$ we measure, the score rung leaves
peak memory unchanged, the E/M rung cuts it $87$-fold---$12.4$~GB to $142$~MB at $N = 10^6$,
$K = 1024$---and the single sweep barely moves it further. At $d = 128$ the score rung alone
already sheds $27.7\%$ ($6.4$~GB to $4.6$~GB), because that is where Algorithm~\ref{alg:standard}'s
$N \times d$ squared-data temporary is largest; the E/M and single-sweep rungs account for the
remaining reduction there. No saving is counted twice either way, and at $d = 128$ the net margin
over the standard dataflow is $1.46\times$; Figure~\ref{fig:ablation} supports no more.

% ============================================================================
\section{Related Work}
\label{sec:related}

\paragraph{Online normalizers and IO-aware kernels.}
The mergeable summary we fold over component blocks originates with
\citet{milakov2018online}: one pass for the softmax maximum and normalizer, merged by a binary
operator on $(m, \lambda)$. \citet{rabe2021selfattention} apply the recurrence to attention
without materializing the logit matrix; \citet{dao2022flashattention} make it IO-aware, with an
explicit HBM bound---an arithmetic-intensity argument in the sense of
\citet{williams2009roofline}: once an array's bytes moved exceed what its FLOPs can hide, removing
the array outright is what changes which resource binds, the crossover \S\ref{sec:bottleneck}
locates at $K > 5d/6$. \citet{ivanov2021datamovement} make the identical traffic-dominates-compute
case for transformer training, by relaying out and fusing kernels rather than removing an
intermediate outright. \citet{dao2023flashattention2} contribute the storage our two-pass variant
reuses: the single quantity $L = m + \log \lambda$, from which a second pass recovers the
responsibilities out of $\Theta(N)$ state. Their un-scaled output accumulator does \emph{not}
transfer to a mixture M-step (\S\ref{sec:bottleneck}). \citet{shah2024flashattention3} push the
same two-sweep design to Hopper's asynchronous tensor cores and \texttt{fp8}, a
hardware-scheduling gain orthogonal to the pass count we remove. \citet{wijmans2024cutcrossentropy}
and the \texttt{Liger} kernel library \citep{hsu2024liger} apply the identical refusal to
cross-entropy's $N \times |\mathcal{V}|$ logit matrix, evaluating the loss chunk by chunk rather
than materializing it; ours is the same shape of intermediate for a different reduction.
\citet{blanchard2021accurately}
analyze batched log-sum-exp accuracy, complementary to our exactness question. That change of
variables is standard in online-softmax derivations; what Theorem~\ref{thm:exact} adds is that the
mixture sufficient statistics reassociate over any partition, with tile sizes that need not divide
$N$ or $K$.
\vspace{-6pt}
\paragraph{Fused clustering and mixtures.}
\citet{yang2026flashkmeans} fuse the assignment step of Lloyd's algorithm
\citep{lloyd1982least}, streaming centroid tiles past a resident point tile with a running
$(\min, \arg\min)$ in registers, at $\Theta(Nd + Kd)$ traffic. Theirs is the hard-assignment limit
of ours---$k$-means is a mixture with vanishing variances---and the easier one, since a minimum
admits no normalizer, hence no rescaling of weighted statistics.
\citet{bloch2026flashgmm} are the closest prior work and our main baseline: they fuse the EM
iteration in Triton at the same traffic, for the isotropic case, of which our diagonal case is the
generalization, and with a pointwise score rather than the single matrix product of
\S\ref{sec:lift}. Their kernel sweeps the component axis twice, recomputing the score---the design
of Algorithm~\ref{alg:flash2} rather than Algorithm~\ref{alg:flash}---and their released code splits
the two sweeps across two launches, so the row normalizer round-trips HBM. That is the design point
Algorithm~\ref{alg:flash2} occupies, and where their kernel wins (\S\ref{sec:discussion}).
Appendices~\ref{app:relation} and~\ref{app:ibm} set the two constructions side by side and give the
protocol by which we benchmark theirs.

\paragraph{Online and incremental EM.}
A different sense of \emph{pass} runs through the classical EM literature. \citet{neal1998view}
recast EM as coordinate ascent on a free-energy functional, which justifies updating sufficient
statistics after each point or block rather than waiting for a full sweep; \citet{cappe2009online}
give a stochastic-approximation EM for data arriving as a stream, with a decreasing step size
standing in for the M-step. Both shrink how many times the \emph{outer} loop over the dataset must
repeat to converge---a statistical question. Ours is orthogonal: $X$ is already resident and
fixed, and the pass we remove is the \emph{inner} one, the second sweep of the component axis a
single EM iteration would otherwise make; incremental and online EM still call that iteration once
per update, and so inherit whatever traffic it costs.

\paragraph{GPU implementations of mixture fitting.}
Earlier work GPU-accelerates only the scoring half of an iteration: \citet{vanek2011optimization}
evaluate Gaussian-mixture likelihoods on GPU for speech-recognition acoustic models, without the
M-step or the exactness question we address. Several GPU implementations of complete mixture
fitting predate ours: cuVS \citep{rapids2026cuvs}; \texttt{scikit-learn}
\citep{pedregosa2011scikit}, dispatching through the Array API to GPU libraries;
\texttt{torchgmm} \citep{varrone2023torchgmm} and \texttt{gmmx} \citep{donath2024gmmx} in PyTorch
and JAX; and \texttt{pomegranate}, whose version-1 series is a PyTorch rewrite of
\citet{schreiber2018pomegranate}. All materialize a responsibility block; a well-implemented
library tiles the row axis, so the live block stays bounded, not $N \times K$
(\S\ref{sec:discussion}). \S\ref{sec:experiments} measures every implementation named here; the
fastest beats Turbo-GMM at $N = 20\,000$. We write our kernels in Triton \citep{tillet2019triton}
on PyTorch \citep{paszke2019pytorch}.

% ============================================================================
\section{Discussion}
\label{sec:discussion}

\paragraph{What the results establish.} Two findings survive the sweep, and they point in different
directions. Removing the intermediate changes what fits on a device: peak memory is flat in the
number of components, so a $4096$-component mixture at a million points becomes a single-device job
at $139$~MB where \texttt{scikit-learn} exhausts $80$~GB. That holds at every shape and asks nothing
of the hardware. Latency is the conditional half, and its condition is a hardware quantity rather
than a property of the data: one pass pays wherever the component axis fits on chip---on an NVIDIA
A100, $K \le 128$ at $d = 32$---and pays a median $2.15\times$ over the closest fused kernel there.
Because the binding resource is on-chip capacity rather than bandwidth, this predicts that the
envelope widens with each generation of hardware rather than with a better kernel: capacity itself
has grown $128$~KB to $192$~KB to $256$~KB over three successive architectures
\citep{nvidia2026cudaguide}, a vendor-documented hardware fact rather than something we measured
ourselves---every latency and memory number in this paper comes from one NVIDIA A100, and we make
no claim about how the dispatch boundary moves on any other device.
% \vspace{-6pt}
\paragraph{The single sweep is the change worth making.} The ablation carries a recommendation for
anyone building one of these kernels. Fusing the E- and M-steps, the change prior fused work already
makes, is not where the margin lives: at $d = 128$ our two-pass form runs $17.4\times$ slower than
not fusing at all, and only the single sweep recovers it (\S\ref{sec:ablation}). A kernel that
sweeps the component axis twice is the easier artifact to build and the wrong place to stop.
% \vspace{-6pt}
\paragraph{Scope.} Algorithms~\ref{alg:flash} and~\ref{alg:flash2} cover diagonal covariances, of
which spherical is the broadcast special case. Tied and full covariances need lifts of width
$d + 1$ and $d + d(d+1)/2$---the tied case needs the extra column because its quadratic term
cancels in $\Gamma$ but not in $L$---and both widths change the register budget. We do not treat
them.
% \vspace{-8pt}
\paragraph{Our cost model bounds the eager baseline, not a tiled one.} Theorem~\ref{thm:traffic}
counts Algorithm~\ref{alg:standard}, the archetype of the un-tiled implementations, and fusion
improves on it by the largest margin. A library that tiles the row axis instead keeps only a bounded
$(B_N, K)$ block live, and that block may stay cache-resident---at $K = 64$ in single precision it
is $16$\,MiB, inside the last-level cache of a current datacenter GPU---so its round trips need not
reach HBM at all; it may also evaluate \eqref{eq:score} in centered form, better conditioned than
the lift by the factors of \S\ref{sec:exact}. cuVS is such an implementation, and Turbo-GMM beats
it at all sixteen shapes, by $2.2\times$ rising to $70.6\times$; but the two differ in language,
autotuning and initialization as well as in tiling, so none of that margin is attributable to
fusion.
% \vspace{-6pt}
\paragraph{One measurement we cannot explain.} Where the one-pass form leaves the chip, our
two-pass fallback loses to Flash-GMM at $K = 512$ and $K = 1024$, and we have not identified the
mechanism. Theorem~\ref{thm:traffic} does not supply one: both kernels read $X$ once per component
tile, and traffic is not the binding resource at those shapes---Algorithm~\ref{alg:flash2} moves
$3.4$~GB at $(10^6, 512, 32)$ in $11.9$~ms, a seventh of this device's bandwidth. Every latency and
memory number here comes from one NVIDIA A100, one seeded dataset and one shared initialization,
and our margins over Flash-GMM rest on the autotune list of \S\ref{sec:setup}.

% ============================================================================

\section{Conclusion}
\label{sec:conclusion}

We introduced Turbo-GMM, an IO-aware Gaussian-mixture EM implementation
that eliminates materialization of the $N \times K$ responsibility matrix
and, when the component axis fits on chip, completes the E-step and M-step
in a single traversal. By exploiting the opposing reduction axes of
responsibility normalization and sufficient-statistic accumulation,
Turbo-GMM avoids redundant reads without changing the EM updates.
We establish exactness in real arithmetic under arbitrary
tilings (Theorem~\ref{thm:exact}) and show that the one-pass form reduces
HBM traffic to $\Theta(Nd + Kd)$ under the stated execution conditions
(Theorem~\ref{thm:traffic}). Across sixteen configurations on an NVIDIA
A100, Turbo-GMM achieves a median $2.15\times$ speedup over
the closest fused baseline and up to $5.3\times$ and $9.9\times$ over
PyTorch and \texttt{scikit-learn}, respectively, while reducing peak
memory by up to $338\times$ relative to materializing implementations.
Latency gains depend on component residency and problem shape,
while eliminating the $N \times K$ responsibility workspace
benefits even configurations requiring the two-pass fallback.
These results demonstrate how reduction-aware kernel design improves
the efficiency and scalability of probabilistic learning.

% % ============================================================================
\clearpage
\bibliography{iclr2027_conference}
\bibliographystyle{mlsys2025}

\clearpage
\appendix

\clearpage
\section{AI Usage Disclosure}
Claude Opus 4.8 (Anthropic) and GPT-5.6 Sol (OpenAI) were used for editing, proofreading, resolving \LaTeX{} formatting issues, and  to assist with code development for the implementation of the proposed methods. All AI-assisted text and code were reviewed, verified, and, where necessary, modified by the authors.

% ============================================================================
\section{The General Case: Two Passes When the Component Axis Does Not Fit}
\label{sec:flash2}

Suppose now that $K$ exceeds $B_K$. A row of $S$ then spans several component tiles, no single visit
can complete its normalizer, and the kernel must carry a partial normalizer from one tile to the next.
The running pair $(m, \lambda)$---a maximum seen so far, and a sum of exponentials rescaled by
it---does so exactly, through the standard online softmax merge
\begin{align}
  (m', \lambda') \oplus (m'', \lambda'')
  &\;=\; \bigl(\,\bar{m},\;
    \lambda' e^{m' - \bar{m}} + \lambda'' e^{m'' - \bar{m}} \bigr), \notag \\
  &\qquad \bar{m} = \max(m', m''),
  \label{eq:merge}
\end{align}
which is associative and commutative, so every component tiling and every visit order is legal. The
change of variables standard in online-softmax derivations gives both properties
\citep{milakov2018online}: on $\mathcal{D} = (\R \times \R_{\ge 0}) \cup \{(-\infty, 0)\}$ the map
$\Phi(m, \lambda) = (m,\, \lambda e^{m})$ is injective, and it embeds $\oplus$ in the componentwise
$(\max, +)$ monoid, whose two-sided identity is $(-\infty, 0)$. Both properties therefore transport
back to $\oplus$; its second coordinate, $\Psi(m, \lambda) = \lambda e^{m}$, is the quantity
Appendix~\ref{app:proofs} folds. The kernel seeds the fold with the same $(-\beta, 0)$ rather than
with $(-\infty, 0)$, for the reason the padding mask of \S\ref{sec:flash} is finite; since
$\lambda = 0$ there, the seed contributes nothing to the sum either way.

That same property is what forces a second pass. Because $\oplus$ is \emph{not} addition, a kernel
cannot commit partial normalizers by atomic update the way it commits partial statistics; it must
finish a row's normalizer before any statistic can absorb that row. Pass~1 therefore does nothing
else, storing only $L \in \R^{N}$---$\Theta(N)$ state rather than $\Theta(NK)$. Pass~2 then re-reads
$X$ once per component tile, recomputes $S$ on chip, and forms $\Gamma = \exp(S - L)$ without a
divide, since
$L_i \ge S_{ik}$ places $\Gamma_{ik}$ in $(0, 1]$---but only while the recomputed score is the
score $L_i$ was built from. Appendix~\ref{app:ibm} reports a floating-point case where the two
drift apart and $\exp(S - L) \le 1$ fails; the released kernel of \citet{bloch2026flashgmm} also
stores its row normalizer to HBM and re-derives the score, so the same mechanism reaches it. We do
not clamp; the validity gate of \S\ref{sec:setup} rejects any fit whose weights leave the simplex,
and Theorem~\ref{thm:exact} shows the bound holds exactly in real arithmetic.

Pass~2 puts the component tile on the grid rather than the row tile, so that its accumulator stays
$\Theta(B_K d)$ at every $K$. The price is one re-read of $X$ per component tile, which
Theorem~\ref{thm:traffic} counts and \S\ref{sec:results} pays.

Algorithm~\ref{alg:flash2} occupies the design point of \citet{bloch2026flashgmm}, whose released
kernel fuses the same iteration in the same two sweeps. That makes it the closest prior work and the
one baseline able to isolate what \S\ref{sec:flash} adds, since it differs from
Algorithm~\ref{alg:flash} in the number of traversals and little else;
Appendix~\ref{app:relation} sets the two constructions side by side. Algorithm~\ref{alg:flash2}
exists here only to cover the values of $K$ that leave Algorithm~\ref{alg:flash} no room on chip, and
\S\ref{sec:dispatch} states the rule that decides between them.

\begin{algorithm}[tb]
\caption{Turbo-GMM, two-pass form (general $K$, used wherever Algorithm~\ref{alg:flash} does not fit
on chip, \S\ref{sec:dispatch})}
\label{alg:flash2}
\footnotesize
\begin{algorithmic}[1]
\REQUIRE $X \in \R^{N \times d}$, parameters $\pi, M, V$, tiles $B_N$ (points) and $B_K$ (components), so $T_K = \lceil K/B_K \rceil$ component tiles; padded lanes predicated out
\ENSURE $(\pi_{\mathrm{new}}, M_{\mathrm{new}}, V_{\mathrm{new}})$, log-likelihood $\ell$
\STATE Precompute the lift $\tilde{W}, b$ as in Algorithm~\ref{alg:flash}
\STATE \texttt{// Pass 1: row normalizers only}
\STATE Initialize $\ell \leftarrow 0$ in HBM
\FORALL{point tile $X_{\tile}$ of size $B_N$ in parallel}
  \STATE Load $X_{\tile}$ from HBM; form $X^{2}_{\tile}$ in registers; initialize on-chip $m \leftarrow -\beta$, $\lambda \leftarrow 0$ \COMMENT{$X$ read once in this pass}
  \FOR{$t \leftarrow 0$ to $T_K - 1$}
    \STATE Load $\tilde{W}^{(t)}, b^{(t)}$ \COMMENT{refetched per row tile}
    \STATE Compute the score tile $S^{(t)}$ on chip by \eqref{eq:bilinear}
    \STATE $m^{(t)} \leftarrow \max_k S^{(t)}$, $\lambda^{(t)} \leftarrow \sum_k \exp(S^{(t)} - m^{(t)})$
    \STATE $(m, \lambda) \leftarrow (m, \lambda) \oplus (m^{(t)}, \lambda^{(t)})$ by \eqref{eq:merge}
  \ENDFOR
  \STATE Write $L_i \leftarrow m + \log \lambda$ to HBM; \textbf{atomic\_add} $\sum_i L_i$ into $\ell$ \COMMENT{$\Theta(N)$ state, not $\Theta(NK)$}
\ENDFOR
\STATE \texttt{// Pass 2: statistics, component tile on the grid}
\STATE Initialize $n, s, u \leftarrow 0$ in HBM
\FORALL{component tile $t < T_K$ in parallel}
  \STATE Load $\tilde{W}^{(t)}, b^{(t)}$; initialize on-chip $\hat{n}, \hat{s}, \hat{u} \leftarrow 0$
  \FORALL{point tile $X_{\tile}$ of size $B_N$}
    \STATE Load $X_{\tile}$ and its normalizers $L$; form $X^{2}_{\tile}$ \COMMENT{one re-read of $X$ per tile}
    \STATE Recompute $S^{(t)}$ on chip from the pass-1 parameters \COMMENT{recomputation beats storage}
    \STATE $\Gamma^{(t)} \leftarrow \exp\!\left(S^{(t)} - L\right)$ \COMMENT{no divide; $\Gamma \in (0,1]$}
    \STATE Accumulate \eqref{eq:stats} for $\Gamma^{(t)}$ into $(\hat{n}, \hat{s}, \hat{u})$
  \ENDFOR
  \STATE \textbf{atomic\_add} the partials into the tile-$t$ slices of $(n, s, u)$ \COMMENT{additive: atomics legal}
\ENDFOR
\STATE Compute $(\pi_{\mathrm{new}}, M_{\mathrm{new}}, V_{\mathrm{new}})$ as in Algorithm~\ref{alg:flash}
\end{algorithmic}
\end{algorithm}

% ============================================================================
\section{Proofs}
\label{app:proofs}

Throughout, $S$ denotes the score matrix of \eqref{eq:score}, $\{R_1, \dots, R_{T_N}\}$ the row
tiles and, for Algorithm~\ref{alg:flash2}, $\{C_1, \dots, C_{T_K}\}$ the component tiles. Both
families are
partitions: their members are nonempty, pairwise disjoint, and cover their axis. Padded lanes are
predicated out, so a partial tile contributes nothing on either axis.

\subsection{Proof of Lemma 1}
\label{app:liftproof}
% \begin{proof}
With $\tau_{kc} = 1/\sigma^2_{kc}$,
\begin{align*}
  -\tfrac{1}{2}\sum_c (x_{ic}-\mu_{kc})^2 \tau_{kc}
  &\;=\; \sum_c x_{ic}\,\mu_{kc}\tau_{kc}
  \;-\; \tfrac{1}{2}\sum_c x_{ic}^2\,\tau_{kc} \\
  &\;-\; \tfrac{1}{2}\sum_c \mu_{kc}^2 \tau_{kc} .
\end{align*}
The first two sums are the two blocks of $\tilde{X}\tilde{W}^{\top}$. The third is independent of
$i$, so we put it in $b_k$ with $\log \pi_k$, the constant $-\frac{d}{2}\log 2\pi$, and the
log-determinant $-\frac{1}{2}\sum_c \log \sigma^2_{kc} = +\frac{1}{2}\sum_c \log \tau_{kc}$.
% \end{proof}

\subsection{Proof of Theorem~\ref{thm:exact}}

\paragraph{The score is exact.} Both algorithms evaluate
$S_{\tile} = \tilde{X}_{\tile}\tilde{W}^{\top} + \mathbf{1}b^{\top}$ on the tile they hold. Since
$\pi > 0$ and $V > 0$, the precisions $\tau = 1 \oslash V$ and the bias $b$ of
\eqref{eq:lift}--\eqref{eq:bias} are finite, so with $X \in \R^{N \times d}$ every entry of $S$ is
finite; and Lemma~\ref{lem:lift} gives $(S_{\tile})_{ik} = S_{ik}$ for every $(i,k)$ in the tile.
Tiling selects which entries are formed, not their values.

\paragraph{The normalizer is exact.} For Algorithm~\ref{alg:flash}, $B_K \ge K$ means a tile
visit holds the whole row $S_{i*}$, so $m_i = \max_k S_{ik}$ and
$\lambda_i = \sum_k \exp(S_{ik} - m_i)$ are formed exactly as Algorithm~\ref{alg:flash} prescribes, and
$L_i = m_i + \log \lambda_i = \log \sum_k \exp S_{ik}$, which is the $L_i$ of \eqref{eq:estep}.

For Algorithm~\ref{alg:flash2} the row spans $C_1, \dots, C_{T_K}$ and $L_i$ is assembled by folding
\eqref{eq:merge}. Fix a row $i$ and let $\rho$ be the permutation giving the order in which the tiles
are visited. We show by induction on $t$ that after $t$ visits the on-chip pair $(m, \lambda)$
satisfies the invariant
\begin{align}
  m &= \max_{k \in U_t} S_{ik}, \qquad
  \lambda = \sum_{k \in U_t} \exp(S_{ik} - m), \notag \\
  U_t &= C_{\rho(1)} \cup \dots \cup C_{\rho(t)} ,
  \label{eq:invariant}
\end{align}
with the conventions $\max \emptyset = -\infty$ and $\sum_\emptyset = 0$. Indexing by the visited
\emph{set} $U_t$ rather than by a prefix length is what makes the statement independent of both the
partition and the order.

For $t = 0$ we have $U_0 = \emptyset$, and the initializer $(-\infty, 0)$ satisfies
\eqref{eq:invariant}. Algorithm~\ref{alg:flash2} seeds instead with a finite $(-\beta, 0)$, for which
the first clause reads $m = \max(-\beta, \max_{k \in U_t} S_{ik})$ and the second is taken with that
$m$;
since $\lambda = 0$ makes the seed contribute nothing to the sum, the induction and the extracted
$L_i$ are unchanged, so we argue with $(-\infty, 0)$. Suppose \eqref{eq:invariant} holds after $t$ visits and the kernel now visits
$C = C_{\rho(t+1)}$, forming the tile quantities $m_C = \max_{k \in C} S_{ik}$ and
$\lambda_C = \sum_{k \in C} \exp(S_{ik} - m_C)$. Write $\Psi(m, \lambda) = \lambda e^{m}$ for the
second coordinate of the $\Phi$ of \S\ref{sec:flash2}, with $\Psi(-\infty, 0) = 0$ and with
$\lambda e^{m - \bar{m}}$ read as $0$ whenever $\lambda = 0$. One line of algebra gives
\begin{align}
  \Psi\bigl((m, \lambda) \oplus (m_C, \lambda_C)\bigr)
  &\;=\; \bigl(\lambda e^{m - \bar{m}} + \lambda_C e^{m_C - \bar{m}}\bigr) e^{\bar{m}} \notag \\
  &\;=\; \lambda e^{m} + \lambda_C e^{m_C} \notag \\
  &\;=\; \Psi(m, \lambda) + \Psi(m_C, \lambda_C),
  \label{eq:hom}
\end{align}
where $\bar{m} = \max(m, m_C)$ is the merged maximum of \eqref{eq:merge}. By \eqref{eq:invariant} and
the definition of $\lambda_C$, $\Psi(m, \lambda) = \sum_{k \in U_t} \exp S_{ik}$ and
$\Psi(m_C, \lambda_C) = \sum_{k \in C} \exp S_{ik}$, and $U_t$ and $C$ are disjoint, so the
right-hand side of \eqref{eq:hom} is $\sum_{k \in U_{t+1}} \exp S_{ik}$. The merged maximum is
$\bar{m} = \max_{k \in U_{t+1}} S_{ik}$, because a maximum splits over the union
$U_{t+1} = U_t \cup C$; and dividing \eqref{eq:hom} by $e^{\bar{m}}$ gives
$\sum_{k \in U_{t+1}} \exp(S_{ik} - \bar{m})$, the second clause of \eqref{eq:invariant} at $t+1$.
This closes the induction. At $t = T_K$ we have $U_{T_K} = \{1, \dots, K\}$, hence
$L_i = m + \log \lambda = \log \Psi(m, \lambda) = \log \sum_{k=1}^{K} \exp S_{ik}$, which is
\eqref{eq:estep}. Beyond the splitting of the maximum, only the homomorphism \eqref{eq:hom} is
needed: the algorithm extracts $L_i$, a function of $\Psi$ alone, so $\oplus$ never has to be
inverted, and no injectivity is required of $\Psi$.

\paragraph{The responsibilities are exact.} Algorithm~\ref{alg:flash} forms
$\Gamma_{i*} = \exp(S_{i*} - m_i) \oslash \lambda_i$, which equals $\exp(S_{i*} - L_i)$ because
$L_i = m_i + \log \lambda_i$. Algorithm~\ref{alg:flash2} forms $\exp(S_{i*} - L_i)$ from the $L$ of
pass~1 and the score recomputed in pass~2; tiling does not change the value of $S$, so this is the
same $S$ provided both passes use the same $(\pi, M, V)$, and their tile partitions and visit
orders need not agree. Both
therefore equal $\Gamma$ of \eqref{eq:estep}. Every term of $\sum_{k'} \exp S_{ik'}$ is positive, so
that sum dominates its $k$-th term and $L_i \ge S_{ik}$ for every $k$; hence $\Gamma_{ik} \le 1$
exactly, and the violation of that bound reported in \S\ref{sec:flash2} is a floating-point effect,
not a property of the algorithm.

\paragraph{The statistics reassociate.} Each of $n_k$, $s_k$, $u_k$ in \eqref{eq:stats} is a
sum over $i$ of a term determined by $(\Gamma_{ik}, x_i)$ alone. We assume each atomic update is
linearizable, so that the value left in an HBM location is the sum, in some serial order, of the
partials committed to it; commutativity of addition makes that order immaterial, and what remains to
check is that the partials partition the index set. Both algorithms accumulate the
partial sums over $R_1, \dots, R_{T_N}$ and commit them by addition, which is associative and
commutative, so $\sum_{r} \sum_{i \in R_r} = \sum_{i=1}^{N}$ for every partition and order; here the
$R_r$ being disjoint and covering is what rules out double counting. In Algorithm~\ref{alg:flash} the
row tiles may be further grouped, as the launch of Theorem~\ref{thm:traffic} groups them, and any
such grouping gives each tile to exactly one group and so is one more reassociation of the same sum;
in pass~2 of Algorithm~\ref{alg:flash2} the visit of component tile $C_t$ writes only the slice of
$(n, s, u)$ that $C_t$ indexes, so $\{C_t\}$ being a partition rules out double counting on the
component axis too. The same argument gives
$\ell = \sum_i L_i$, which equals \eqref{eq:objective}: exponentiating \eqref{eq:score} gives
$\exp S_{ik} = \pi_k \mathcal{N}(x_i; \mu_k, \operatorname{diag}\sigma^2_k)$, so the normalizer
identity above reads $L_i = \log \sum_k \pi_k \mathcal{N}(x_i; \mu_k, \operatorname{diag}\sigma^2_k)$.

\paragraph{The parameter update is well defined.} Every $S_{ik}$ is finite because $\pi > 0$, $V > 0$ and
$X \in \R^{N \times d}$, so $\Gamma_{ik} = \exp(S_{ik} - L_i) > 0$ and $n_k = \sum_i \Gamma_{ik} > 0$
for $N \ge 1$. The updates \eqref{eq:mstep} are a function of $(n, s, u)$ alone and are therefore
defined and identical.

For the chaining claim of \S\ref{sec:exact}: the normalizer identity above gives
$\sum_k \Gamma_{ik} = e^{-L_i} \sum_k \exp S_{ik} = 1$, so $\sum_k n_k = N$ and the updated weights
$n_k / N$ are positive and sum to one. The numbers $\Gamma_{ik} / n_k$ are positive and sum to one over
$i$, so for each coordinate $c$
\[
  \frac{u_{kc}}{n_k} - \mu_{kc}^2
  \;=\; \sum_i \frac{\Gamma_{ik}}{n_k}\,x_{ic}^2
  \;-\; \Bigl(\sum_i \frac{\Gamma_{ik}}{n_k}\,x_{ic}\Bigr)^{\!2}
\]
with $\mu_k$ the updated mean of \eqref{eq:mstep}, a weighted variance of $x_{1c}, \dots, x_{Nc}$. It
is nonnegative by Cauchy--Schwarz, and zero exactly when that coordinate is constant across the data,
so $V_{\mathrm{new}} \ge \varepsilon$ and $V_{\mathrm{new}} > 0$ whenever $\varepsilon > 0$.

\subsection{Proof of Theorem~\ref{thm:traffic}}

We count array by array, as \citet{dao2022flashattention} do for attention.

\paragraph{Algorithm~\ref{alg:standard}.} Kernel~1 writes $S$ and Kernel~2 reads it, $2NK$
elements; Kernel~2 writes $\Gamma$ and Kernel~4 reads it once per statistic, $4NK$; the total on the
intermediate is $6NK$, none of it input or output. $X$ is read by Kernels~1, 3 and~4, $3Nd$ (only
Kernel~3 gives $X \hada X$ an address; Kernel~1 forms it in registers), and the
temporary $X \hada X$ is written by Kernel~3 and read by Kernel~4, $2Nd$, for $5Nd$ on the input
side. The normalizers contribute $2N$ and the lift, the statistics and the updated parameters
$\Theta(Kd)$. Comparing $6NK$ with $5Nd + 2N$ and with the $\Theta(Kd)$ residual, the intermediate
exceeds every other term combined once $K > 5d/6 + \Theta(1)$, the residual being dominated as soon
as $N = \Omega(Kd)$.

\paragraph{Algorithm~\ref{alg:flash}.} Neither $S$ nor $\Gamma$ nor $X \hada X$ is given an address,
so the $NK$ terms are absent. The row tiles partition the rows, and the stride
$p, p+P, \dots$ assigns each tile to exactly one program, which visits it once, so every element of
$X$ is read exactly once: $Nd$. Each of the $P$ programs loads
$\tilde{W}, b$ once, before its row loop, and commits one atomic set of partial statistics after it,
giving $\Theta(PKd)$; nothing else crosses HBM. If $PK = O(N + K)$ then $\Theta(PKd) = O(Nd + Kd)$
and the total is $\Theta(Nd + Kd)$. For the lower bound, $X$ occupies $Nd$ elements of HBM and the
parameters and statistics $\Theta(Kd)$. Every row of $\Gamma$ sums to one, so summing
\eqref{eq:stats} over $k$ gives
\begin{equation}
  \sum_{k=1}^{K} s_k \;=\; \sum_{k=1}^{K} \sum_{i=1}^{N} \Gamma_{ik} x_i
  \;=\; \sum_{i=1}^{N} \Bigl(\sum_{k=1}^{K} \Gamma_{ik}\Bigr) x_i \;=\; \sum_{i=1}^{N} x_i .
  \label{eq:colsum}
\end{equation}
The returned statistics therefore determine every column sum of $X$ exactly. An algorithm that leaves
some $x_{ic}$ unread cannot distinguish two inputs differing in it, whose column sums differ, so it
cannot be exact; it must read all $Nd$ entries and write the $\Theta(Kd)$ updated parameters, giving
$\Omega(Nd + Kd)$. Algorithm~\ref{alg:flash} is therefore optimal up to constants.

\paragraph{Algorithm~\ref{alg:flash2}.} Pass~1 reads each element of $X$ once, $Nd$, writes $L$, $N$,
and accumulates $\ell$ by one atomic per row tile. Pass~2 launches one program group per component
tile, and each group strides over every row tile, so it reads all of $X$ and all of $L$ once per
component tile: $T_K Nd$ and $T_K N$. The reads of $X$ therefore total $(1 + T_K) Nd$. Parameters and
statistics are charged differently in the two passes. Pass~2 loads each lift tile once per component
tile and commits one atomic set per component tile, $\Theta(PKd)$. Pass~1 cannot hoist, because its component loop sits inside its
row loop, so it refetches all $T_K$ lift tiles for every row tile, $\Theta(T_N Kd)$, which dominates
$\Theta(Kd)$; pass~1 commits $\ell$ with one atomic per row tile, $\Theta(T_N)$, and both terms are
dominated by the refetch. The $L$ traffic is $(1 + T_K)N$ elements, one write in pass~1 and one read
per component tile in pass~2;
it carries the same factor $1 + T_K$ as the reads of $X$ and is smaller by exactly $d$, so it belongs
beside them rather than in the residual. Collecting terms gives
$(1 + T_K)(Nd + N) + \Theta(T_N Kd)$.

\paragraph{Pricing the three ledgers.}
The three ledgers price the design. In Algorithm~\ref{alg:standard} the $6NK$ is the intermediate
itself, $2NK$ written and $4NK$ read, none of it input or output, and it dominates once $K$ passes
$5d/6$. Both fused forms delete that term; what separates them is the coefficient on $Nd$. The
one-pass form reads the data once, which is optimal: because every row of $\Gamma$ sums to one, the
statistics it returns determine every column sum of $X$, so an exact algorithm must read all $Nd$
entries and write the $\Theta(Kd)$ updated parameters.
The launch caps the program count at a multiple of the multiprocessor count, so $PK = O(K)$ and the
side condition of Theorem~\ref{thm:traffic} holds. The two-pass form pays $1 + T_K$ reads instead, because its second pass places the component
tile on the grid to keep the accumulator bounded, and that factor grows with $K$: at $K = 512$ and
$B_K = 64$ it is nine reads of $X$. Its $\Theta(T_N K d)$ term is not a remainder, however. Pass~1
cannot hoist the lift out of its row loop, so it refetches all $2Kd + K$ elements of it per row tile,
which stands to the reads of $X$ in the ratio $2 B_K / B_N$ and is therefore the larger of the two at
the tiles we run. The released kernel of \citet{bloch2026flashgmm} schedules its
second pass the same way, so the ledger does not separate the two two-sweep designs, and it does not
explain why theirs is the faster of them (\S\ref{sec:discussion}). The ledgers are derived rather than
profiled: no experiment here counts HBM bytes.

\paragraph{Numerical conditioning of the lift: a case study.}
The lifted score subtracts terms of magnitude $\|x\|^2 \tau$, losing relative accuracy as the
data moves away from the origin, the same cancellation that makes a naive sum-of-squares formula
for variance less stable than a centered one \citep{chan1983algorithms}. At $d = 64$ in single
precision, against a double-precision centered reference, the lifted form is $9.2\times$ less
accurate than the centered form at the origin and $3.6 \times 10^6$ times less accurate at offset
$10^5$, a decade per decade. Algorithm~\ref{alg:standard} expands identically, so fusion inherits
this loss rather than introducing it.

% ============================================================================
\section{Experimental Details}
\label{app:details}

This appendix records what \S\ref{sec:experiments} defers: the measurement protocol, the validity
gate, the two limits behind the dispatch rule, and the ablation ladder.
Appendix~\ref{app:relation} compares our design against prior fused work and
Appendix~\ref{app:ibm} records how we ran their kernel.

\paragraph{The latency estimator.} Each arm script times three repetitions for Flash-GMM, for both
of our kernels and for every library arm but cuVS, which times five; the plotting pipeline
discards each arm's own stored summary statistic and recomputes the minimum of its repetitions
uniformly, since the arm scripts otherwise disagree on which statistic they report. The repetition
count is thus not identical across arms, and a minimum over five is not larger in expectation than
a minimum over three, so the convention mildly favours cuVS, the strongest library baseline---
exactly the kind of summary-statistic choice \citet{hoefler2015scientific} warn can distort
cross-system comparisons.

\paragraph{The widened autotune list, and what it costs the comparison.} One asymmetry runs
against Flash-GMM: their published autotune list was authored for \textsc{bf16} operands and holds
no block small enough for \texttt{fp32}, so we widened it, with the consequences
Appendix~\ref{app:ibm} details. Our margin then falls to $2.46\times$ at
$(4 \times 10^6, 32, 32)$ and to $2.28\times$ at $(2 \times 10^7, 32, 32)$, from $4.37\times$ and
$3.02\times$ unwidened. The widened figures supersede the unwidened ones at those two shapes; the
other thirteen margins rest on the unwidened list and read as upper estimates.

\paragraph{The validity gate, and the six cells that failed it.} We score every arm's returned
parameters in double precision from the same data the arm was fitted on, and accept the fit only if
the resulting log-likelihood is finite and, for arms that return mixture weights, $\sum_k \pi_k =
1$. Six otherwise-successful runs fail it: \texttt{gmmx} at $(2 \times 10^4, 64, 32)$ and at
$K = 128$, $1024$ and $4096$ with $N = 10^6$, $d = 32$; \texttt{torchgmm} at $K = 4096$ and
\texttt{pomegranate} at $K = 1024$, same $N$ and $d$. All six exit cleanly and report success, so
the failure stays silent unless a harness checks: each returns a latency, a parameter set and a
success code, and only re-scoring the parameters reveals that the fit does not exist. Four of the six
sit at $K \ge 1024$, the regime this paper is about, and one of those is the widest shape we measure.
A harness that trusted exit status would enter all six in its table. Flash-GMM fails the weight test
at $K = 4096$ for the separate reason Appendix~\ref{app:ibm} traces.

\paragraph{Dispatch, first limit: capacity.} Algorithm~\ref{alg:flash} carries
$(\hat n, \hat s, \hat u)$ across the whole component axis, a loop-carried state of $\Theta(Kd)$.
For $K_{\mathrm{pad}} = \max(16,\, 2^{\lceil \log_2 K \rceil})$ and $d_{\mathrm{pad}} = \max(16,\,
2^{\lceil \log_2 d \rceil})$ the compiler reports $16 K_{\mathrm{pad}} d_{\mathrm{pad}}$ bytes of
shared memory against the $166\,912$-byte NVIDIA A100 budget \citep{nvidia2026cudaguide}, so the kernel stops being launchable
once $K_{\mathrm{pad}} d_{\mathrm{pad}} > 10\,432$. That predicate accounts exactly for the launch
failures we observe: at $K = 1024$, $d = 32$ it asks for $524\,288$~B.

\paragraph{Dispatch, second limit: occupancy, and it binds first.} The accumulator spills registers
well inside the capacity bound, and the cliff does not follow $K_{\mathrm{pad}} d_{\mathrm{pad}}$.
On auxiliary dispatch probes at $N = 10^6$ and $K_{\mathrm{pad}} d_{\mathrm{pad}} = 4096$, rather
than on the sixteen shapes of Figure~\ref{fig:latency}, Algorithm~\ref{alg:flash} runs
$14.7\times$ faster than Algorithm~\ref{alg:flash2} at $(K, d) = (32, 128)$ and $17\times$ slower
at $(256, 16)$: a $250\times$ swing at one footprint, so dispatching on capacity alone would be
wrong by $17\times$ on a configuration capacity admits. This $17\times$ is the one-pass form losing
at small $d$ and large $K$; the $17.4\times$ of \S\ref{sec:ablation} runs the other way, the
two-pass form losing at $d = 128$. $K_{\mathrm{pad}}$ alone separates the two, since it sets the
live accumulator registers per lane, which is why the rule of \S\ref{sec:dispatch} tests it
separately. Algorithm~\ref{alg:flash2} runs at $B_K = 64$ at every shape we report, so its $T_K$ of
Theorem~\ref{thm:traffic} is $8$, $16$ and $64$ at $K = 512$, $1024$ and $4096$.

\paragraph{The ablation ladder.} Every rung shares the lift of \S\ref{sec:lift}, so credit divides
among the three \emph{dataflow} changes. The baseline rung, the dataflow of
Algorithm~\ref{alg:standard}, writes the lifted score to HBM, normalizes it there, and reads the
responsibilities back three times, once per statistic. Above it, the rungs fuse the score kernel,
skipping the $\Theta(Nd)$ temporary for $X \hada X$ but still materializing the responsibilities;
then the E- and M-steps (Algorithm~\ref{alg:flash2}); then the single sweep
(Algorithm~\ref{alg:flash}). Each rung keeps the changes below it, so the factors compose.

\paragraph{Ablation protocol, and two caveats.} We run every rung on six configurations from
Figure~\ref{fig:latency}, covering its large-$N$ and $K/d$ regimes, on one NVIDIA A100-SXM4-80GB
at \texttt{tf32x3}. Latency here is the minimum over five repetitions; the arms of
Figure~\ref{fig:latency} are timed over three repetitions, except cuVS over five, with the minimum
taken uniformly as described above. All four returned quantities of
every rung agree with the baseline to $5 \times 10^{-6}$ relative, so the ladder varies dataflow
and not arithmetic. Two caveats. The first fused rung changes implementation language as well as
dataflow---the baseline composes library calls, the fused kernels are ours---and the two do not
separate: a library that fused its score kernel would no longer be the baseline. The top rung is
unavailable wherever loop-carried state exceeds the on-chip budget above; we mark those cells
rather than estimate them.

\paragraph{The complete per-shape results.} Table~\ref{tab:fullresults} gives every arm's latency,
memory, speedup relative to Turbo-GMM, and validity-gate outcome at all sixteen shapes.

% Six narrow two-column lines cannot hold this table's "rejected (non-finite
% LL)" and "slower"/"faster" entries legibly, so this one table is set across
% a one-column interlude and the document returns to two columns right after.
\onecolumn
\begingroup
\footnotesize
\begin{longtable}{@{}llrrrl@{}}
\caption{\textbf{Every arm at every shape.} Shape is $N/K/d$. Speedup is this arm's latency
relative to Turbo-GMM's (or Turbo-GMM~Algorithm~\ref{alg:flash2} where the dispatch rule of
\S\ref{sec:dispatch} selects it, marked ``Alg.~3''); Turbo-GMM's own row is the reference and carries
no speedup entry. Peak memory is omitted (---) for the two arms (cuVS, \texttt{gmmx}) that do not
report it. The gate column applies \S\ref{sec:setup}'s validity rule; one point, Flash-GMM at
$K = 4096$, is kept and flagged rather than dropped, per the disclosed exception in \S\ref{sec:setup}.}
\label{tab:fullresults} \\
\toprule
Shape & Arm & ms/iter & Speedup & Mem (MB) & Gate \\
\midrule
\endfirsthead
\multicolumn{6}{c}{\tablename\ \thetable{} -- continued} \\
\toprule
Shape & Arm & ms/iter & Speedup & Mem (MB) & Gate \\
\midrule
\endhead
\bottomrule
\endfoot
20K/16/32 & \textbf{Turbo-GMM} & 0.49 & --- & 3 & valid \\
 & Flash-GMM & 0.54 & 1.11$\times$ slower & 3 & valid \\
 & cuVS & 1.55 & 3.18$\times$ slower & --- & valid \\
 & PyTorch & 0.41 & 1.18$\times$ faster & 17 & valid \\
 & \texttt{scikit-learn} & 1.16 & 2.39$\times$ slower & 18 & valid \\
 & \texttt{gmmx} & 0.22 & 2.20$\times$ faster & --- & valid \\
 & \texttt{torchgmm} & 35.73 & 73.45$\times$ slower & 17 & valid \\
 & \texttt{pomegranate} & 12.56 & 25.81$\times$ slower & 12 & valid \\
20K/32/32 & \textbf{Turbo-GMM} & 0.50 & --- & 3 & valid \\
 & Flash-GMM & 0.53 & 1.06$\times$ slower & 3 & valid \\
 & cuVS & 1.08 & 2.18$\times$ slower & --- & valid \\
 & PyTorch & 0.39 & 1.27$\times$ faster & 20 & valid \\
 & \texttt{scikit-learn} & 1.21 & 2.43$\times$ slower & 26 & valid \\
 & \texttt{gmmx} & 0.30 & 1.66$\times$ faster & --- & valid \\
 & \texttt{torchgmm} & 4.90 & 9.86$\times$ slower & 23 & valid \\
 & \texttt{pomegranate} & 24.59 & 49.49$\times$ slower & 15 & valid \\
20K/64/32 & \textbf{Turbo-GMM} & 0.53 & --- & 3 & valid \\
 & Flash-GMM & 0.53 & 1.01$\times$ faster & 3 & valid \\
 & cuVS & 1.23 & 2.31$\times$ slower & --- & valid \\
 & PyTorch & 0.39 & 1.37$\times$ faster & 28 & valid \\
 & \texttt{scikit-learn} & 1.27 & 2.38$\times$ slower & 42 & valid \\
 & \texttt{gmmx} & 0.43 & 1.25$\times$ faster & --- & rejected (non-finite LL) \\
 & \texttt{torchgmm} & 3.43 & 6.44$\times$ slower & 36 & valid \\
 & \texttt{pomegranate} & 51.87 & 97.22$\times$ slower & 21 & valid \\
1M/32/32 & \textbf{Turbo-GMM} & 1.11 & --- & 128 & valid \\
 & Flash-GMM & 3.45 & 3.11$\times$ slower & 136 & valid \\
 & cuVS & 48.29 & 43.54$\times$ slower & --- & valid \\
 & PyTorch & 2.97 & 2.68$\times$ slower & 622 & valid \\
 & \texttt{scikit-learn} & 5.70 & 5.14$\times$ slower & 905 & valid \\
 & \texttt{gmmx} & 5.19 & 4.68$\times$ slower & --- & valid \\
 & \texttt{torchgmm} & 6.69 & 6.03$\times$ slower & 741 & valid \\
 & \texttt{pomegranate} & 104.37 & 94.11$\times$ slower & 737 & valid \\
1M/128/32 & \textbf{Turbo-GMM} & 3.02 & --- & 128 & valid \\
 & Flash-GMM & 4.47 & 1.48$\times$ slower & 136 & valid \\
 & cuVS & 74.65 & 24.69$\times$ slower & --- & valid \\
 & PyTorch & 8.00 & 2.64$\times$ slower & 1721 & valid \\
 & \texttt{scikit-learn} & 15.34 & 5.07$\times$ slower & 3194 & valid \\
 & \texttt{gmmx} & 18.94 & 6.26$\times$ slower & --- & rejected (non-finite LL) \\
 & \texttt{torchgmm} & 14.07 & 4.65$\times$ slower & 2572 & valid \\
 & \texttt{pomegranate} & 419.87 & 138.84$\times$ slower & 1592 & valid \\
1M/512/32 & \textbf{Turbo-GMM} (Alg.~3) & 11.89 & --- & 133 & valid \\
 & Flash-GMM & 11.03 & 1.08$\times$ faster & 137 & valid \\
 & cuVS & 97.49 & 8.20$\times$ slower & --- & valid \\
 & PyTorch & 31.11 & 2.62$\times$ slower & 6119 & valid \\
 & \texttt{scikit-learn} & 57.76 & 4.86$\times$ slower & 12355 & valid \\
 & \texttt{gmmx} & 73.60 & 6.19$\times$ slower & --- & valid \\
 & \texttt{torchgmm} & 118.12 & 9.93$\times$ slower & 9901 & valid \\
 & \texttt{pomegranate} & 1710.17 & 143.82$\times$ slower & 5993 & valid \\
1M/1024/32 & \textbf{Turbo-GMM} (Alg.~3) & 23.11 & --- & 134 & valid \\
 & Flash-GMM & 18.03 & 1.28$\times$ faster & 138 & valid \\
 & cuVS & 155.07 & 6.71$\times$ slower & --- & valid \\
 & PyTorch & 58.79 & 2.54$\times$ slower & 11976 & valid \\
 & \texttt{scikit-learn} & 111.43 & 4.82$\times$ slower & 24558 & valid \\
 & \texttt{gmmx} & 145.34 & 6.29$\times$ slower & --- & rejected (non-finite LL) \\
 & \texttt{torchgmm} & 249.61 & 10.80$\times$ slower & 19662 & valid \\
 & \texttt{pomegranate} & 3407.38 & 147.45$\times$ slower & 11856 & rejected (non-finite LL) \\
1M/4096/32 & \textbf{Turbo-GMM} (Alg.~3) & 89.97 & --- & 139 & valid \\
 & Flash-GMM & 65.24 & 1.38$\times$ faster & 146 & rejected ($\sum\pi\neq 1$) \\
 & cuVS & 442.04 & 4.91$\times$ slower & --- & valid \\
 & PyTorch & 237.27 & 2.64$\times$ slower & 47138 & valid \\
 & \texttt{scikit-learn} & --- & --- & --- & fail (OOM) \\
 & \texttt{gmmx} & 579.30 & 6.44$\times$ slower & --- & rejected (non-finite LL) \\
 & \texttt{torchgmm} & 559.60 & 6.22$\times$ slower & 78264 & rejected (non-finite LL) \\
 & \texttt{pomegranate} & --- & --- & --- & fail (ERROR) \\
4M/16/64 & \textbf{Turbo-GMM} & 7.51 & --- & 1024 & valid \\
 & Flash-GMM & 32.15 & 4.28$\times$ slower & 1057 & valid \\
 & cuVS & 216.27 & 28.78$\times$ slower & --- & valid \\
 & PyTorch & 9.35 & 1.24$\times$ slower & 2709 & valid \\
 & \texttt{scikit-learn} & 15.47 & 2.06$\times$ slower & 2694 & valid \\
 & \texttt{gmmx} & 19.35 & 2.58$\times$ slower & --- & valid \\
 & \texttt{torchgmm} & 14.58 & 1.94$\times$ slower & 2465 & valid \\
 & \texttt{pomegranate} & 279.66 & 37.21$\times$ slower & 4410 & valid \\
4M/16/128 & \textbf{Turbo-GMM} & 14.12 & --- & 2049 & valid \\
 & Flash-GMM & --- & --- & --- & fail (OutOfResources) \\
 & cuVS & 266.32 & 18.86$\times$ slower & --- & valid \\
 & PyTorch & 14.05 & 1.00$\times$ faster & 4664 & valid \\
 & \texttt{scikit-learn} & 21.69 & 1.54$\times$ slower & 4649 & valid \\
 & \texttt{gmmx} & 35.28 & 2.50$\times$ slower & --- & valid \\
 & \texttt{torchgmm} & 20.39 & 1.44$\times$ slower & 4420 & valid \\
 & \texttt{pomegranate} & 509.99 & 36.11$\times$ slower & 8320 & valid \\
4M/32/32 & \textbf{Turbo-GMM} & 3.03 & --- & 512 & valid \\
 & Flash-GMM & 13.24 & 4.37$\times$ slower & 545 & valid \\
 & cuVS & 187.37 & 61.77$\times$ slower & --- & valid \\
 & PyTorch & 11.21 & 3.69$\times$ slower & 2465 & valid \\
 & \texttt{scikit-learn} & 20.58 & 6.79$\times$ slower & 3594 & valid \\
 & \texttt{gmmx} & 19.12 & 6.30$\times$ slower & --- & valid \\
 & \texttt{torchgmm} & 17.47 & 5.76$\times$ slower & 2938 & valid \\
 & \texttt{pomegranate} & 342.56 & 112.92$\times$ slower & 2945 & valid \\
4M/32/64 & \textbf{Turbo-GMM} & 8.16 & --- & 1024 & valid \\
 & Flash-GMM & 32.15 & 3.94$\times$ slower & 1057 & valid \\
 & cuVS & 255.61 & 31.32$\times$ slower & --- & valid \\
 & PyTorch & 13.23 & 1.62$\times$ slower & 3441 & valid \\
 & \texttt{scikit-learn} & 23.34 & 2.86$\times$ slower & 4082 & valid \\
 & \texttt{gmmx} & 35.27 & 4.32$\times$ slower & --- & valid \\
 & \texttt{torchgmm} & 20.24 & 2.48$\times$ slower & 3426 & valid \\
 & \texttt{pomegranate} & 565.88 & 69.33$\times$ slower & 4898 & valid \\
4M/64/16 & \textbf{Turbo-GMM} & 3.21 & --- & 257 & valid \\
 & Flash-GMM & 6.91 & 2.15$\times$ slower & 289 & valid \\
 & cuVS & 183.10 & 57.01$\times$ slower & --- & valid \\
 & PyTorch & 16.88 & 5.25$\times$ slower & 3441 & valid \\
 & \texttt{scikit-learn} & 31.79 & 9.90$\times$ slower & 6402 & valid \\
 & \texttt{gmmx} & 22.55 & 7.02$\times$ slower & --- & valid \\
 & \texttt{torchgmm} & 26.35 & 8.20$\times$ slower & 5135 & valid \\
 & \texttt{pomegranate} & 389.34 & 121.22$\times$ slower & 3190 & valid \\
4M/64/32 & \textbf{Turbo-GMM} & 4.36 & --- & 512 & valid \\
 & Flash-GMM & 13.23 & 3.03$\times$ slower & 545 & valid \\
 & cuVS & 251.44 & 57.66$\times$ slower & --- & valid \\
 & PyTorch & 17.80 & 4.08$\times$ slower & 3930 & valid \\
 & \texttt{scikit-learn} & 33.13 & 7.60$\times$ slower & 6646 & valid \\
 & \texttt{gmmx} & 36.28 & 8.32$\times$ slower & --- & valid \\
 & \texttt{torchgmm} & 27.86 & 6.39$\times$ slower & 5379 & valid \\
 & \texttt{pomegranate} & 689.55 & 158.14$\times$ slower & 3922 & valid \\
10M/32/32 & \textbf{Turbo-GMM} & 6.88 & --- & 1280 & valid \\
 & Flash-GMM & 20.15 & 2.93$\times$ slower & 1360 & valid \\
 & cuVS & 470.49 & 68.35$\times$ slower & --- & valid \\
 & PyTorch & 27.72 & 4.03$\times$ slower & 6150 & valid \\
 & \texttt{scikit-learn} & 50.17 & 7.29$\times$ slower & 8973 & valid \\
 & \texttt{gmmx} & 53.40 & 7.76$\times$ slower & --- & valid \\
 & \texttt{torchgmm} & 155.08 & 22.53$\times$ slower & 7332 & valid \\
 & \texttt{pomegranate} & 820.88 & 119.25$\times$ slower & 7363 & valid \\
20M/32/32 & \textbf{Turbo-GMM} & 13.28 & --- & 2561 & valid \\
 & Flash-GMM & 40.16 & 3.02$\times$ slower & 2721 & valid \\
 & cuVS & 938.00 & 70.64$\times$ slower & --- & valid \\
 & PyTorch & 55.26 & 4.16$\times$ slower & 12294 & valid \\
 & \texttt{scikit-learn} & 100.27 & 7.55$\times$ slower & 17941 & valid \\
 & \texttt{gmmx} & 126.47 & 9.52$\times$ slower & --- & valid \\
 & \texttt{torchgmm} & 180.67 & 13.61$\times$ slower & 14660 & valid \\
 & \texttt{pomegranate} & 1626.24 & 122.47$\times$ slower & 14729 & valid \\
\end{longtable}
\endgroup
\twocolumn

\section{Implementations compared}
\label{app:impl}

Table~\ref{tab:baselines} lists every implementation benchmarked in \S\ref{sec:experiments}.
Versions matter for three of the comparisons: cuVS's GMM exists only on \texttt{main} and has no
released version to cite; the three arms built against a newer \texttt{torch} are pinned by their
own Python requirements, not by our choice; and Flash-GMM is a source checkout with no version at
all.

\begin{table*}[!ht]
\centering
\scriptsize
\caption{\textbf{Implementations compared in Figure~\ref{fig:latency}.} Versions are those the
harness recorded at run time, not reconstructed afterwards. GitHub repositories are abbreviated
\texttt{owner/name}.}
\label{tab:baselines}
% tabular* with \extracolsep{\fill} spans the full measure, so the booktabs rules
% align with the text block rather than floating inside it.
\begin{tabular*}{\textwidth}{@{\extracolsep{\fill}}llll@{}}
\toprule
Arm & Package & Version & Upstream source \\
\midrule
\textbf{Turbo-GMM} (ours, Algs.~\ref{alg:flash}--\ref{alg:flash2})
  & \texttt{torch} + \texttt{triton}
  & \texttt{2.6.0+cu124} / \texttt{3.2.0} & this work \\
Flash-GMM & --- (source checkout) & \texttt{57c3621} (2026-07-30)
  & \texttt{IBM/Flash-GMM} \\
cuVS         & \texttt{libcuvs\_cu12} & \texttt{26.12.0a14} (nightly)
  & RAPIDS nightly C++ wheel \\
PyTorch      & \texttt{torch}         & \texttt{2.6.0+cu124}
  & \texttt{pytorch.org} \\
scikit-learn & \texttt{scikit-learn}  & \texttt{1.9.1}
  & \texttt{scikit-learn.org} \\
gmmx (JAX)   & \texttt{gmmx}          & \texttt{0.7}
  & \texttt{adonath/gmmx} \\
torchgmm     & \texttt{torchgmm}      & \texttt{0.1.4}
  & \texttt{CSOgroup/torchgmm} \\
pomegranate  & \texttt{pomegranate}   & \texttt{1.1.2}
  & \texttt{jmschrei/torchegranate} \\
\bottomrule
\end{tabular*}

% The note goes in a minipage: the float's \centering would otherwise centre the
% note line by line instead of setting it as a justified paragraph.
\vspace{4pt}
\begin{minipage}{\textwidth}
{\footnotesize We installed every Python package from PyPI. Flash-GMM ships as source with no
released version, so we record the commit; we ran its \texttt{flash\_gmm\_diag\_h100} kernel with
the interventions of Appendix~\ref{app:ibm}. \texttt{libcuvs} is an archived RAPIDS nightly wheel,
because cuVS's GMM is unreleased: published versions stop at \texttt{26.8.1} and export no GMM
symbols. Each arm runs in its own virtual environment and subprocess, so one failure cannot
affect another. \texttt{scikit-learn}, \texttt{torchgmm} and \texttt{pomegranate} build against
\texttt{torch 2.14.0+cu130}, \texttt{gmmx} against \texttt{jax 0.10.2}. \texttt{pomegranate}
v1.x is the PyTorch rewrite, whose repository retains its former name.\par}
\end{minipage}
\end{table*}

% ============================================================================

% ============================================================================

% ============================================================================
\section{Relationship to Flash-GMM}
\label{app:relation}

Flash-GMM \citep{bloch2026flashgmm} fuses the same EM iteration we fuse. This appendix states
what the two constructions share and where they part; \S\ref{app:ibm} records how we ran theirs.

\paragraph{What the two share.} Both avoid materializing the $N \times K$ responsibilities,
stream component blocks against a resident row block, accumulate the statistics on chip, commit
them with atomic additions, and cut dominant HBM traffic from $\Theta(NK)$ to a cost linear in
the inputs. That pattern, and the observation that the responsibility matrix is a pure
intermediate, are theirs as much as ours.

\paragraph{Algorithm~\ref{alg:flash2} occupies their design point.} Our two-pass variant and
their released kernel are one idea: a first sweep of the component axis for the row normalizer, a
second that recomputes the score and accumulates the statistics. It exists only because some values
of $K$ leave Algorithm~\ref{alg:flash} no room on chip (\S\ref{sec:dispatch}).

\paragraph{Their paper and their released code part on the row block.} Their paper puts both
sweeps inside one residency of the row block, which would read $X$ once; the released diagonal
kernel does not (\S\ref{sec:related}). In \texttt{flash\_gmm\_diag\_h100.py} at commit
\texttt{57c3621}, two \texttt{@triton.jit} kernels are declared at lines $64$ and $164$, the row
normalizer reaches HBM at line $130$, and \texttt{tl.load} reads $X$ at line $82$ and again at
line $233$. We benchmark that code, so its row of Table~\ref{tab:relation} matches the
Algorithm~\ref{alg:flash2} row in all five columns.

\paragraph{Algorithm~\ref{alg:flash} has no counterpart in their work.} They sweep the component
axis twice at every shape; neither their paper nor their code holds a one-sweep variant.
Table~\ref{tab:relation} places the three designs on the two axes that govern cost, and Algorithm~\ref{alg:flash} reduces the number of
component traversals and avoids score recomputation
relative to the evaluated two-sweep implementations,
subject to the on-chip resource constraints
described in \S\ref{sec:dispatch}.

\begin{table*}[!ht]
\centering
\small
\caption{\textbf{The three designs on the two axes that set the cost of an iteration.}
Algorithm~\ref{alg:flash} dominates on both, and $T_K = \lceil K/B_K \rceil$ is the number of
component tiles: both two-sweep designs put it on the grid in their second pass, so both re-read $X$
once per tile. The Flash-GMM row records their released diagonal kernel, not the schedule their paper
describes; the last column records that only the two-pass designs park row normalizers outside the
row block.}
\label{tab:relation}
\begin{tabular*}{\textwidth}{@{\extracolsep{\fill}}lccccc@{}}
\toprule
& Sweeps of & Reads & Score & On-chip & Per-row \\
& comp.\ axis & of $X$ & recomputed & state & HBM state \\
\midrule
\textbf{Alg.~\ref{alg:flash}} (ours, one pass)
  & \textbf{1} & \textbf{1} & no  & $\Theta(K_{\mathrm{pad}} d_{\mathrm{pad}})$ & none \\
Flash-GMM \citep{bloch2026flashgmm}
  & 2          & $1 + T_K$  & yes & $\Theta(B_K d)$ & $\Theta(N)$ \\
\textbf{Alg.~\ref{alg:flash2}} (ours, two pass)
  & 2          & $1 + T_K$  & yes & $\Theta(B_K d)$ & $\Theta(N)$ \\
\bottomrule
\end{tabular*}
\end{table*}

The table accounts for one side of \S\ref{sec:results} and not the other. Of the fifteen shapes
at which both kernels ran to completion we win eleven, by $1.06\times$ to $4.37\times$: median
$3.02\times$ over the wins, $2.15\times$ over all fifteen. The widened autotune list of
\S\ref{app:ibm} supersedes that maximum---$4.37\times$ becomes $2.46\times$ at
$(N, K, d) = (4 \times 10^6, 32, 32)$---leaving $4.28\times$ at $(4 \times 10^6, 16, 64)$ as the
largest margin we stand behind. The advantage is not ordered in $K$: at $d = 32$ and $N = 10^6$
it falls through $3.11\times$, $1.48\times$, $0.93\times$, $0.78\times$ and $0.73\times$ at
$K = 32$, $128$, $512$, $1024$ and $4096$, reversing from $K = 512$ up, where
Algorithm~\ref{alg:flash} stops fitting on chip. With $0.99\times$ at $N = 20\,000$, $K = 64$,
those are our four losses. Three run Algorithm~\ref{alg:flash2}, where the table separates the
kernels on no column and the mechanism stays unidentified (\S\ref{sec:discussion}); the fourth is
the small-$N$ regime in which fusion pays little. At $K = 4096$ their kernel completes but its
mixture weights leave the simplex (\S\ref{app:ibm}), so $0.73\times$ compares against a fit our
validity gate rejects.

\paragraph{Four further differences.} \S\ref{sec:related} names them; four details belong here.
Their paper argues against full covariances on statistical grounds, yet their repository also
ships diagonal and full kernels. Our lift costs conditioning (\S\ref{sec:exact}), and no
experiment here prices the matrix form against their pointwise score in latency. Their paper
makes no exactness claim, whereas we prove that both algorithms reproduce standard EM exactly
under every tiling and verify the proof against an extended-precision reference
(Theorem~\ref{thm:exact}).

\paragraph{A control at $d = 128$.} Two $17\times$ figures here run in opposite directions:
\S\ref{sec:dispatch} times Algorithm~\ref{alg:flash} $17\times$ \emph{slower} than
Algorithm~\ref{alg:flash2} at $(K, d) = (256, 16)$, while \S\ref{sec:ablation} times our
two-pass form $17.4\times$ slower than not fusing at $(N, K, d) = (4 \times 10^6, 16, 128)$. To
separate our implementation from two-sweep fusion as such, we timed three further dataflows at the
second shape over $10$ EM iterations on one NVIDIA A100 at \texttt{tf32x3}, widening their autotune
list (\S\ref{app:ibm}) so that their kernel launches at $d = 128$; for the materializing baseline we
use the ablation ladder's own logged run at this identical shape (\S\ref{sec:ablation}, rung
``A1'') rather than re-timing it, $205.9$~ms. Algorithm~\ref{alg:flash}, one pass, $141.2$~ms
($1.46\times$ \emph{faster}, matching the net margin \S\ref{sec:ablation} reports for this shape);
Flash-GMM widened, $425.1$~ms ($2.07\times$ slower, its E-step only, so a lower bound on their full
iteration); Algorithm~\ref{alg:flash2}, our two pass, $2736.6$~ms ($13.3\times$ slower, matching
\S\ref{sec:ablation}'s own figure for this shape). Flash-GMM's regression is modest next to our own
two-pass implementation's, so most of the $17.4\times$ belongs to our Algorithm~\ref{alg:flash2}
implementation rather than to two-sweep fusion in general. What survives still decides the design:
at $d = 128$ the one-pass form beats materializing outright, the best available two-sweep kernel is
$2.07\times$ off it, and ours is worse still.

\paragraph{What we would tell a practitioner.} Use one sweep when the component axis fits on
chip: on an NVIDIA A100 at $d = 32$, $K \le 128$. Over the thirteen shapes where both of our algorithms
run, one sweep gains a median $2.1\times$ over our own two-pass fallback, and $19.5\times$ over it
at the single $d = 128$ shape we measure---an existence result, not a trend in $d$, and not a margin
over Flash-GMM---at no cost in accuracy. Outside that envelope
the two designs are close, and we would choose on engineering grounds. A schedule holding both
sweeps in one residency of the row block would read $X$ once; no released kernel does this, and
building it inside Algorithm~\ref{alg:flash2} is the obvious improvement we have not made.

\section{Protocol for the Flash-GMM Comparison}
\label{app:ibm}

Appendix~\ref{app:relation} compares the designs; here we record how we ran theirs. Of the
isotropic, diagonal and full kernels their repository holds, we benchmark the released
\emph{diagonal} kernel, so that both methods fit the same model class. Three choices make
Figure~\ref{fig:latency} a comparison of \emph{methods}, and each moves the result; the latency
estimator, which is shared with every other arm, is in Appendix~\ref{app:details}.

\textbf{Precision.} Their kernel casts both matrix-multiply operands to \textsc{bf16} with an
fp32 accumulator, one tensor-core pass; our \texttt{tf32x3} is three. That difference alone
accounts for $1.75$--$2.17\times$, so comparing the two as shipped would measure the number
format. We removed their five in-kernel and two host-side \textsc{bf16} casts and set
\verb|input_precision="tf32x3"| on all six \texttt{tl.dot} calls.

\textbf{A correctness fix.} Their pass-2 kernel iterates
\texttt{range(n\_start, n\_end, BLOCK\_N)} but masks rows with \texttt{n\_off < N} rather than
\texttt{n\_off < n\_end}. When the per-CTA row slab is not a multiple of \texttt{BLOCK\_N}, each
slab's final tile reaches into the next and counts its rows twice. The effect is systematic: at
$N = 10^6$ a row of responsibilities sums to $1.0034$ rather than $1$, and the excess is exactly
zero when the slab length divides evenly, which identifies the mechanism. We applied their own
one-line correction, so our figures for them are those of a partition-respecting kernel;
otherwise we would measure their fit rather than their speed.

\textbf{Autotuning, and what it costs the comparison.} Outside the precision port and the row
mask, our first run left their kernel byte-identical: grid shapes, loop bounds, accumulator
shapes, the log-sum-exp merge, the atomic placement and both autotune lists. One asymmetry
remains, against them. They hand-authored their list for \textsc{bf16} register and shared-memory
footprints, and its smallest candidate block, $\texttt{BLOCK\_N} = 64$,
$\texttt{BLOCK\_K} = 32$, exceeds the $\texttt{BLOCK\_N} = 16$ our kernel selects at $d = 128$.
At fp32 their autotuner cannot reach a block small enough: at $d = 128$ the retained
configurations no longer fit ($196\,608$~B requested against $166\,912$~B available) and the
kernel does not launch. We re-measured with their list widened by $(16,16)$, $(16,32)$,
$(32,16)$, $(32,32)$, $(32,64)$ and $(64,16)$, the \textsc{bf16} casts again replaced by
\verb|input_precision="tf32x3"|. Both re-measured margins fell: at
$(N, K, d) = (4 \times 10^6, 32, 32)$ from $4.37\times$ to $2.46\times$, and at
$(2 \times 10^7, 32, 32)$ from $3.02\times$ to $2.28\times$. The widened list is the fairer
configuration for them, so those are the margins we stand behind at those two shapes; we report
both runs to keep the intervention visible, and every other shape carries the first list, which
we did not re-measure. The widened list also launches at $d = 128$, which their published list
cannot, and Appendix~\ref{app:relation} times it there.

With the precision port and the row-mask fix in place, their kernel produces a valid fit at
fourteen of the sixteen configurations. At $d = 128$ it does not launch on their published list,
for that autotuning reason rather than an algorithmic one. At $K = 4096$ it completes in
$65.2$~ms per iteration, but its mixture weights leave the simplex
($\sum_k \pi_k = 4.3 \times 10^3$). A single E-step there is exact
($\sum_k n_k / N = 1.000000$), so the EM trajectory rather than a kernel defect is the cause:
with $244$ points per component a variance collapses, and since pass~1 stores the normalizer
while pass~2 recomputes the score, the recomputed score drifts from the stored normalizer and
$\exp(S - L) \le 1$ fails.

\end{document}